\documentclass[sigconf]{acmart}
\usepackage{graphicx}
\usepackage{xcolor}
\usepackage{multirow}
\usepackage{tabularx}
\usepackage{algorithm}
\usepackage{algorithmic}
\usepackage{subcaption}
\usepackage{amsthm}
\usepackage{enumitem}

\newcolumntype{C}{>{\centering\arraybackslash}X} % centered version of 'X' col. type
\newtheorem{problem}{Problem}

\newcommand{{\method}}{RM-GIB}

\copyrightyear{2023}
\acmYear{2023}
\setcopyright{acmlicensed}\acmConference[KDD '23]{Proceedings of the 29th
ACM SIGKDD Conference on Knowledge Discovery and Data Mining}{August
6--10, 2023}{Long Beach, CA, USA}
\acmBooktitle{Proceedings of the 29th ACM SIGKDD Conference on Knowledge
Discovery and Data Mining (KDD '23), August 6--10, 2023, Long Beach, CA,
USA}
\acmPrice{15.00}
\acmDOI{10.1145/3580305.3599248}
\acmISBN{979-8-4007-0103-0/23/08}
\begin{document}
\title{A Unified Framework of Graph Information Bottleneck for Robustness and Membership Privacy}

\author{Enyan Dai}
\email{emd5759@psu.edu}
\affiliation{
\institution{The Pennsylvania State University}
\country{USA}}

\author{Limeng Cui}
\email{culimeng@amazon.com}
\affiliation{
\institution{Amazon}
\country{USA}}

\author{Zhengyang Wang}
\email{zhengywa@amazon.com}
\affiliation{
\institution{Amazon}
\country{USA}}

\author{Xianfeng Tang}
\email{xianft@amazon.com}
\affiliation{
\institution{Amazon}
\country{USA}}

\author{Yinghan Wang}
\email{yinghanw@amazon.com}
\affiliation{
\institution{Amazon}
\country{USA}}

\author{Monica Cheng}
\email{chengxc@amazon.com}
\affiliation{
\institution{Amazon}
\country{USA}}

\author{Bing Yin}
\email{alexbyin@amazon.com}
\affiliation{
\institution{Amazon}
\country{USA}}

\author{Suhang Wang}
\email{szw494@psu.edu}
\affiliation{
\institution{The Pennsylvania State University}
\country{USA}}

%\fancyhead{}
\renewcommand{\shortauthors}{Enyan Dai et al.}
\begin{CCSXML}
<ccs2012>
   <concept>
       <concept_id>10010147.10010257</concept_id>
       <concept_desc>Computing methodologies~Machine learning</concept_desc>
       <concept_significance>500</concept_significance>
       </concept>
 </ccs2012>
\end{CCSXML}

\ccsdesc[500]{Computing methodologies~Machine learning}
\keywords{Graph Neural Networks; Membership Privacy; Robustness}
\begin{abstract}

Graph Neural Networks (GNNs) have achieved great success in modeling graph-structured data. However, recent works show that GNNs are vulnerable to adversarial attacks which can fool the GNN model to make desired predictions of the attacker. In addition, training data of GNNs can be leaked under membership inference attacks. This largely hinders the adoption of GNNs in high-stake domains such as e-commerce, finance and bioinformatics. Though  investigations have been made in conducting robust predictions and protecting membership privacy, they generally fail to simultaneously consider the robustness and membership privacy. Therefore, in this work, we study a novel problem of developing robust and membership privacy-preserving GNNs. Our analysis shows that Information Bottleneck (IB) can help filter out noisy information and regularize the predictions on labeled samples, which can benefit robustness and membership privacy. However, structural noises and lack of labels in node classification challenge the deployment of IB on graph-structured data. To mitigate these issues, we propose a novel graph information bottleneck framework that can alleviate structural noises with neighbor bottleneck. Pseudo labels are also incorporated in the optimization to minimize the gap between the predictions on the labeled set and unlabeled set for membership privacy. Extensive experiments on real-world datasets demonstrate that our method can give robust predictions and simultaneously preserve membership privacy.
\end{abstract}

% \vskip -4em
\maketitle

\section{Introduction}
Graph Neural Networks (GNNs) have shown promising results in modeling graph-structured data  such as social network analysis~\cite{hamilton2017inductive}, finance~\cite{wang2019semi}, and drug discovery~\cite{irwin2012zinc}. For graphs, both graph topology and node attributes are important for downstream tasks. Generally, GNNs adopt a message-passing mechanism to update a node's representation by aggregating information from its neighbors.  The learned node representation can preserve both node attributes and local structural information, which facilitates various tasks, especially semi-supervised node classification.

Despite their great success in modeling graphs, GNNs are at risk of \textit{adversarial attacks} and \textit{privacy attacks}. \textit{First}, GNNs are vulnerable to adversarial attacks~\cite{zugner2019adversarial,zou2021tdgia,dai2018adversarial}. An attacker can achieve various attack goals such as controlling predictions of target nodes~\cite{dai2018adversarial} and degrading the overall performance~\cite{zugner2019adversarial} by deliberately perturbing the graph structure and/or node attributes. For example, Nettack~\cite{zugner2018adversarial} can mislead the target GNN to give wrong predictions on target nodes by poisoning the training graph with small perturbations on graph structure or node attributes. The vulnerability of GNNs largely hinders the adoption of GNNs in safety-critical domains such as finance and healthcare.
\textit{Second}, GNNs might leak private training data information under membership inference attacks (MIAs)~\cite{shokri2015privacy,olatunji2021membership}. The membership inference attack can detect whether a target sample belongs to the training set. It can effectively distinguish the training samples even with black-box access to the prediction vectors of the target GNNs. This potential membership leakage threatens the privacy of the GNN models trained on sensitive data such as clinical records. For example, an attacker can infer the patient list from GNN-based chronic disease prediction on the patient network~\cite{lu2021weighted}.

Many efforts~\cite{zhu2019robust,entezari2020all,jin2020graph,tang2020transferring,zhang2020gnnguard,dai2022comprehensive} have been taken to learn robust GNNs against adversarial attacks. For instance, robust aggregation mechanisms~\cite{zhu2019robust,chen2021understanding,geisler2021robustness,liu2021elastic} have been investigated to reduce the negative effects of adversarial perturbations. A group of graph denoising methods~\cite{zhang2020gnnguard,jin2020graph,entezari2020all,dai2022towards} is also proposed to remove/down-weight the adversarial edges injected by the attacker. Though they are effective in defending graph adversarial attacks, these methods may fail to preserve the membership privacy, which is also empirically verified in Sec.~\ref{sec:exp_adv}. For membership privacy-preserving, approaches such as adversarial regularization~\cite{nasr2018machine} and differential privacy~\cite{abadi2016deep,papernot2016semi} are proposed for independent and identically distributed (i.i.d) data. However, in semi-supervised node classification, the size of labeled nodes is small and information on labeled nodes can be propagated to their neighbor nodes. These will challenge existing methods that generally process i.i.d data with sufficient labels. Work in membership privacy-preserving on GNNs is still limited~\cite{olatunji2021membership}, let alone robust and membership privacy-preserving GNNs. Therefore, in this paper, we focus on a novel problem of simultaneously defending adversarial attacks and membership privacy attacks with a unified framework.

One promising direction of simultaneously achieving robustness and membership privacy-preserving is to adopt the information bottleneck (IB) principle~\cite{tishby2000information} for node classification of GNNs. 
The IB principle aims to learn a code that maximally expresses the target task while containing minimal redundant information. In the objective function of IB, apart from the classification loss, a regularization is applied to constrain information irrelevant to the classification task in the bottleneck code. \textit{First}, as IB encourages filtering out information irrelevant to the classification task, the noisy information from adversarial perturbations could be reduced, resulting in robust predictions~\cite{alemi2016deep}. 
\textit{Second}, membership inference attack is feasible because of the difference between training and test samples in posteriors. As analyzed in Sec~\ref{sec:pre_privacy}, the regularization in IB can constrain the mutual information between representations and labels on the training set, which can narrow the gap between training and test sets to avoid membership privacy leakage. 

Though promising, there are still two challenges in applying IB principle for robust and membership privacy-preserving predictions on graphs. \textit{First}, in graph-structured data, adversarial perturbations can happen in both node attributes and graph structures. However, IB for i.i.d data is only designed to extract compressed information from attributes. Simply extending the IB objective function used for i.i.d data to the GNN model may fail to filter out the structural noises. This problem is also empirically verified in Sec.~\ref{Sec:pre_robust}. \textit{Second}, in semi-supervised node classification, the size of labeled nodes is small. Without enough labels, the IB framework would have poor performance on test nodes. In this situation, the gap between labeled nodes and unlabeled test nodes can still be large even with the IB regularization term on labeled nodes, making it ineffective to defend MIA. Our empirical analysis in Sec.~\ref{sec:pre_privacy} also proves that this challenge is caused by lacking labels.

In an attempt to address these challenges, we propose a novel \underline{R}obust and \underline{M}embership Privacy-Preserving \underline{G}raph \underline{I}nformation \underline{B}ottleneck ({\method}). 
{\method} develops a novel graph information bottleneck framework that adopts an attribute bottleneck and a neighbor bottleneck, which can handle the redundant information and adversarial perturbations in both node attributes and graph topology. Moreover, a novel self-supervisor is deployed to benefit the neighbor bottleneck in alleviating noisy neighbors to further improve the robustness. Since membership privacy-preserving with IB requires a large number of labels, {\method} collects pseudo labels on unlabeled nodes and combines them with provided labels in the optimization to guarantee membership privacy. In summary, our main contributions are:
\begin{itemize}[leftmargin=*]
    \item We investigate a new problem of developing a robust and membership privacy-preserving framework for graphs.
    \item We propose a novel {\method} that can alleviate both attribute and structural noises with bottleneck and preserve the membership privacy through incorporating pseudo labels in the optimization.
    \item Extensive experiments in various real-world datasets demonstrate the effectiveness of our proposed {\method} in defending membership inference and adversarial attacks.
\end{itemize}

\section{Related Works}

\subsection{Graph Neural Networks}
Graph Neural Networks (GNNs)~\cite{kipf2016semi,velivckovic2017graph,ying2018graph,bongini2021molecular} have shown remarkable ability in modeling graph-structured data, which benefits various applications such as recommendation system~\cite{ying2018graph}, drug discovery~\cite{bongini2021molecular} and traffic analysis~\cite{zhao2020semi}. Generally, GNNs adopt a message-passing mechanism to iteratively aggregate the neighbor information to augment the representation learning of center nodes. 
For instance, in each layer of GCN~\cite{kipf2016semi}, the representations of neighbors and the center node will be averaged, followed by a non-linear transformation such as ReLU. GAT~\cite{velivckovic2017graph} deploys an attention mechanism in the neighbor aggregation to benefit the representation learning. 
Recently, many extensions and improvements have been made to address various challenges in graph learning~\cite{dai2021say,qiu2020gcc,dai2021nrgnn}. For example, new frameworks of GNNs such LW-GCN~\cite{dai2022labelwise} are designed to handle the graph with heterophily.
FairGNN~\cite{dai2021say} is proposed to mitigate the bias of predictions of GNNs.
Various self-supervised GNNs~\cite{qiu2020gcc,dai2021towards} have been explored to learn better representations. However, despite the great achievements, GNNs are vulnerable to adversarial~\cite{zugner2018adversarial} and privacy attacks~\cite{olatunji2021membership}, which largely constrain the applications of GNNs in safety-critical domains such as bioinformatics and finance.

\subsection{Robust Graph Learning}
Extensive studies~\cite{wu2019adversarial,zugner2018adversarial,zugner2019adversarial,dai2023unnoticeable} have shown that GNNs are vulnerable to adversarial attacks. Attackers can inject a small number of adversarial perturbations on graph structures and/or node attributes for their attack goals such as reducing overall performance~\cite{zugner2019adversarial,zou2021tdgia} or controlling predictions of target nodes~\cite{zugner2018adversarial,dai2023unnoticeable}. 

Recently, many efforts have been taken to defend against adversarial attacks~\cite{zhu2019robust,entezari2020all,jin2020graph,tang2020transferring,dai2022comprehensive}, which can be roughly divided into three categories, i.e., adversarial training, robust aggregation, and graph denoising. In adversarial training~\cite{xu2019topology}, the GNN model is forced to give similar predictions for a clean sample and its adversarially perturbed version to achieve robustness. The robust aggregation methods~\cite{zhu2019robust,geisler2021robustness,liu2021elastic} design a new message-passing mechanism to restrict the negative effects of adversarial perturbations. Some efforts in adopting Gaussian distributions as hidden representations~\cite{zhu2019robust}, aggregating the median value of each neighbor embedding dimension~\cite{geisler2021robustness}, and incorporating $l_1$-based graph smoothing~\cite{liu2021elastic}. In graph denoising methods~\cite{jin2020graph,wu2019adversarial,entezari2020all,dai2022towards,li2022reliable}, researchers propose various methods to identify and remove/down-weight the adversarial edges injected by the attacker.
For example, {Wu et al.}~\cite{wu2019adversarial} propose to prune the perturbed edges based on the Jaccard similarity of node features. 
% Another preprocessing method by low-rank approximation of adjacent matrix is investigated~\cite{entezari2020all}.
Pro-GNN~\cite{jin2020graph} learns a clean graph structure by low-rank constraint. RS-GNN~\cite{dai2022towards} introduces a feature similarity weighted edge-reconstruction loss to train the link predictor which can down-weight the noisy edges and predict the missing links. However, these methods do not consider defense against membership inference attacks; On the contrary, the proposed {\method} can simultaneously defend against both adversarial attacks and membership inference attacks.

\subsection{Membership Privacy Preservation}
% Apart from adversarial attacks, deep neural networks are also threatened by the privacy attacks~\cite{shokri2015privacy,olatunji2021membership,hu2022membership}. 
Membership inference attack (MIA)~\cite{shokri2015privacy,olatunji2021membership} is a type of privacy attack that aims to identify whether a sample belongs to the training set. The main idea of MIA is to learn a binary classifier on patterns such as posteriors that training and test samples exhibit different distributions. The membership leakage will largely threaten the privacy of the model trained on sensitive data such as medical records.
Many studies~\cite{shokri2017membership,hayes2017logan,choquette2021label,nasr2018machine,abadi2016deep} have been conducted to defend against the membership inference attack on models trained on i.i.d data. The overfitting on the training samples leads to the difference between training samples and test samples in terms of posteriors and other patterns, which makes the membership inference attack feasible. Hence, a group of MIA defense methods propose to reduce the generalization gap through various regularization techniques. For example, L2 regularization~\cite{shokri2017membership}, weight normalization~\cite{hayes2017logan}, and dropout~\cite{salem2018ml,choquette2021label} have been investigated for membership privacy preservation. Adversarial regularization~\cite{nasr2018machine} is also explored to reduce the posterior distribution difference between training and test samples. Another type of defense~\cite{abadi2016deep,chaudhuri2011differentially,papernot2016semi} is to apply differentially private mechanisms such as DP-SGD~\cite{abadi2016deep}. These mechanisms generally add noise to gradients, model parameters, or outputs to achieve membership privacy guarantee. The above membership inference attack and defense methods are mainly on i.i.d data.

Recently, several seminal works~\cite{olatunji2021membership,he2021node,wu2021adapting} show that GNNs also suffer from MIA. However, defending MIA on graphs is rarely explored~\cite{olatunji2021membership}. \citeauthor{olatunji2021membership}~\cite{olatunji2021membership} propose to inject noise to the posteriors or sample neighbors in the aggregation to protect the membership privacy on node classification. However, it will largely sacrifice the node classification performance to achieve membership privacy. On the contrary, our method combines the proposed novel graph IB and pseudo labels to give accurate and membership privacy-preserving predictions. Moreover, our framework is robust to both MIA and adversarial attacks.
% However, the aforementioned methods are overwhelmingly dedicated to the privacy of i.i.d data, which may not be applicable to graphs because the information propagation 

% The leakage of labeled samples 

\subsection{Information Bottleneck}
The Information Bottleneck (IB) principle~\cite{tishby2000information} aims to learn latent representations of each sample that maximally express the target task while containing minimal redundant  information. \citeauthor{alemi2016deep}~\cite{alemi2016deep} firstly propose the variational information bottleneck (VIB) to introduce the IB principle to deep learning. 
As IB filters out information irrelevant to the downstream task, it naturally leads to more robust representations, which have been investigated in~\cite{wang2021revisiting,kim2021distilling,alemi2016deep} for i.i.d data.  
% Since then IB is applied to facilitate the representation learning in various applications such as  image attribution~\cite{}, robust representation learning~\cite{}, and natural language processing~\cite{}. 
\citeauthor{wu2020graph}~\cite{wu2020graph} extend the IB principle to learn robust representations on graph-structured data. IB is also applied to extract informative but compressed subgraphs for graph classification~\cite{sun2022graph,yu2020graph} and graph explanation~\cite{miao2022interpretable}. Our method is inherently different from these methods because: (i) we conduct the first attempt to design a novel IB-based framework for membership privacy-preserving on graph neural networks; (ii) we propose a unified framework that can simultaneously defend against adversarial and membership inference attacks.
\section{Preliminaries}
\subsection{Notations} \label{sec:notation}
% We use upper-case letters (e.g., $X$ and $Y$) to denote random variables and lower-case letters (e.g., $\mathbf{x}$ and $y$) to indicate the realizations of these variables. 
We use $\mathcal{G}=(\mathcal{V},\mathcal{E},\mathbf{X})$ to denote an attributed graph, where $\mathcal{V}=\{v_1,...,v_N\}$ is the set of nodes, $\mathcal{E} \in \mathcal{V} \times \mathcal{V}$ is the set of edges, and $\mathbf{X}=\{\mathbf{x}_1,...,\mathbf{x}_N\}$ is node attribute matrix with $\mathbf{x}_i$ being the node attribute vector of $v_i$. $\mathbf{A} \in \mathbb{R}^{N \times N}$ denotes the adjacency matrix of $\mathcal{G}$, where $\mathbf{A}_{ij}=1$ if $(v_i, v_j) \in \mathcal{E}$ and  $\mathbf{A}_{ij}=0$ otherwise. In this work, we focus on semi-supervised node classification. Only a small set of nodes $\mathcal{V}_L$ are provided with labels $\mathcal{Y}_L=\{y_1,\dots, y_l\}$. $\mathcal{V}_U=\mathcal{V}-\mathcal{V}_L$ denotes the unlabeled nodes. Note that the topology and attributes of $\mathcal{G}$ could contain adversarial perturbations or inherent noises. 
% \suhang{depending on if we conduct both transductive and inductive settings, we might want to explain a little bit}

\subsection{Membership Inference Attack} \label{sec:mia}
\textbf{Attacker's Goal}. The goal of MIA is to identify if a target node was used for training the target model $f_T$ for node classification.

\noindent \textbf{Attacker's Knowledge}. We focus on the defense against black-box membership inference attacks as black-box MIA is a practical setting that is widely adopted in existing MIA methods. Specifically, the attacker can have black-box access to the target model $f_T$ to obtain prediction vectors of queried samples. And a shadow graph dataset $\mathcal{G}_S$ from the same distribution of the graph for training $f_T$ is assumed to be available for the attacker. It can be a subgraph or overlap with the training graph $\mathcal{G}$.

\noindent \textbf{General Framework of MIAs}.
Shadow training~\cite{shokri2017membership,olatunji2021membership} is generally used to train the attack model $f_A$ for MIA. In the shadow training, part of nodes in the shadow dataset, i.e., $\mathcal{V}_S^{in} \subset \mathcal{G}_S$, are used to train a shadow model $f_S$ for node classification to mimic the behaviors of the target model $f_T$. 
% \begin{equation}\small
%     \min_{\theta_S}  \sum_{\mathcal{G}_i \in \mathcal{D}_{S}^{train}} l(f_{S}(\mathcal{G}_i), f_T(\mathcal{G}_i)),
% \end{equation}
% For the shadow model, the training samples and test samples are available for the attacker, which can help to construct a dataset combining 
Then, the attacker can construct a dataset by combining the prediction vectors and corresponding ground truth of membership for the attack model training. Specifically, each  node $v_i \in \mathcal{V}_S^{in}$ used to train $f_S$ is labeled as $1$ (membership) and each node $v_j \in \mathcal{V}_S^{out}$ is labeled as $0$ (non-membership), where $\mathcal{V}_S^{out} = \mathcal{V}_S - \mathcal{V}_S^{out}$. Then, the training process of $f_A$ is formally written as follows:
\begin{equation}%\small
    \min_{\theta_A} - \sum_{v_i \in \mathcal{V}_S^{in}} \log (f_A(\mathbf{\hat{y}}_i^S)) - \sum_{v_i \in \mathcal{V}_S^{out}} \log (1-f_A(\mathbf{\hat{y}}_i^S))
\end{equation}
where $f_A$ denotes the attack model, which is a binary classifier to judge if a node is in the training set or not.  
$\theta_A$ represents the parameters of $f_A$. $\mathbf{\hat y}_i^S$ denotes the prediction vector of node $v_i$ from the shadow model $f_S$. As machine learning model generally overfits on the labeled samples, it is feasible to have a well-trained attack model.
With the trained attack model $f_A$, the membership of a target node $v_t$ can be inferred by $f_A(\hat{\bf y}_t^T)$, where $\hat{\bf y}_t^T$ denotes the prediction vector of $v_t$ given by the target model $f_T$. 

\subsection{Problem Definition}
With the notations in Sec.~\ref{sec:notation} and the description of membership inference attacks in Sec.~\ref{sec:mia}, the problem of learning a robust and membership privacy-preserving GNN can be formally defined as:
\begin{problem}
Given a graph $\mathcal{G}=(\mathcal{V}, \mathcal{E}, \mathbf{X})$ with a small set of nodes $\mathcal{V}_L$ labeled, and edge set $\mathcal{E}$ and attributes $\mathbf{X}$ may be poisoned by adversarial perturbations, we aim to learn a robust and membership privacy-preserving GNN $f_{\mathcal{G}}: \mathcal{G} \rightarrow \mathcal{Y}$ that maintains high prediction accuracy on the unlabeled set $\mathcal{V}_U$ and is resistant to membership inference attacks.%and give prediction vectors on $\mathcal{V}_L$ and $\mathcal{V}_U$ that are resistant to the membership inference attack.
\end{problem}
% \suhang{discuss with Enyan, shall we explain that we also do inductive setting}

\subsection{Preliminaries of Information Bottleneck}
The objective of information bottleneck on i.i.d data is to learn a bottleneck representation $\mathbf{z}=f_{\theta}(\mathbf{x})$ that (i) maximizes the mutual information with label $y$; and (ii) filters out information not related to the label $y$. Various functions can be adopted for $f_\theta$ such as neural networks. 
Formally, the objective function of IB can be written as:
\begin{equation}
    \min_{\theta} -I(\mathbf{z};y) + \beta I(\mathbf{z};\mathbf{x}),
    \label{eq:IB}
\end{equation}
where the former term aims to maximize the mutual information between the bottleneck $\mathbf{z}$ and the label $y$. The latter term constrains the mutual information between $\mathbf{z}$ and input $\mathbf{x}$ to help filter out the redundant information for the classification task. $\beta$ is the Lagrangian parameter that balances two terms.

\subsection{Impacts of IB to Membership Privacy} \label{sec:pre_privacy}
% The objective function of IB in Eq.(\ref{eq:IB}) can be viewed as a combination of a loss for classification and a regularization term. For the regularization on 
As shown in Eq.(\ref{eq:IB}), IB will constrain  $I(\mathbf{z};\mathbf{x})$ on the training set. Based on mutual information properties and the fact that $\mathbf{z}$ is only obtained from $\mathbf{x}$, we can derive the following equation:
\begin{equation} 
\begin{aligned}
I(\mathbf{z};\mathbf{x}) & = I(\mathbf{z};y)+I(\mathbf{z};\mathbf{x}|y)-I(\mathbf{z};y|\mathbf{x}) \\
        & = I(\mathbf{z};y)+I(\mathbf{z};\mathbf{x}|y) \geq I(\mathbf{z};y)
\end{aligned}
\label{eq:pre_mem}
\end{equation}
The details of the derivation can be found in the Appendix~\ref{app:proof}.
The constraint on $I(\mathbf{z};\mathbf{x})$ in the IB objective will simultaneously bound the mutual information $I(\mathbf{z};y)$ on the training set $\mathcal{V}_L$. On the contrary,  classifier without using IB will maximize $I(\mathbf{z};y)$ on the training set $\mathcal{V}_L$ without any constraint. Hence, compared to classifier without using IB regularization, classifier using IB objective is expected to exhibit a smaller gap between the training set and test set. As a result, the member inference attack on classifier trained with IB regularization will be less effective. 
% Then, the gap between the training and test set will likely be reduced. As  the membership inference attack is effective due to the overfitting of the target model, the information bottleneck mechanism can be helpful in preserving membership privacy. 
However, in semi-supervised node classification, only a small portion of nodes are labeled. 
$I(\mathbf{z};y)$ will be only maximized on the small set of labeled nodes $\mathcal{V}_L$. Due to the lack of labels, the performance on unlabeled nodes could be poor. And $I(\mathbf{z};y)$ on unlabeled nodes can still be very low. As a result, even with a constraint on $I(\mathbf{z};y)$, the gap between labeled nodes and unlabeled nodes can still be large when the size of labeled nodes is small, resulting in membership privacy leakage. 

\begin{figure}[t]
    \small
    \centering
    \begin{subfigure}{0.49\linewidth}
        \includegraphics[width=0.88\linewidth]{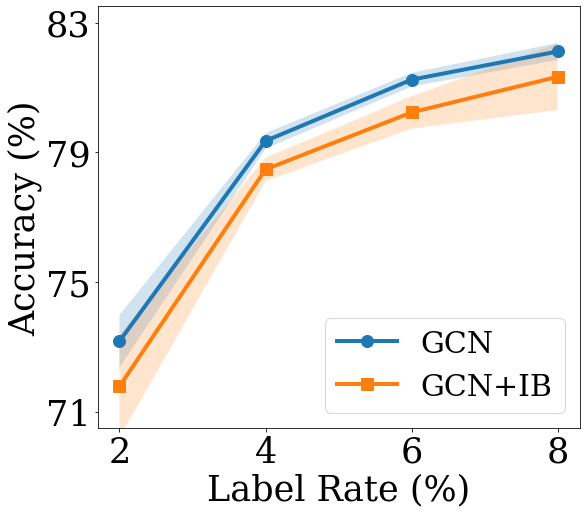}
        \vskip -0.5em
        \caption{Accuracy}
    \end{subfigure}
    \begin{subfigure}{0.49\linewidth}
        \includegraphics[width=0.88\linewidth]{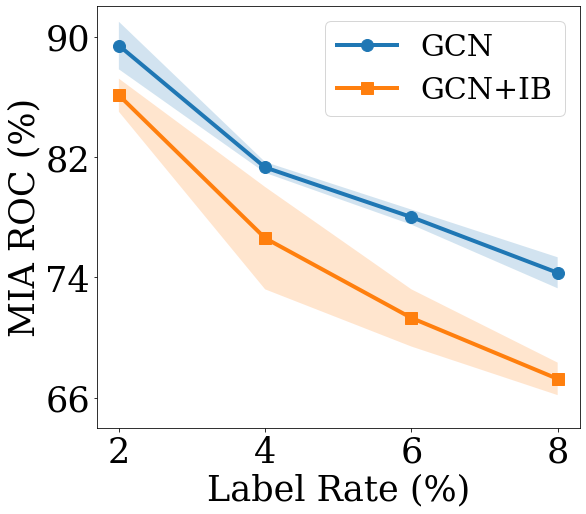}
        \vskip -0.5em
        \caption{MIA-F ROC}
    \end{subfigure}
    \vskip -1em
    \caption{Results of classification and MIA on Cora.}
    \vskip -1.5em
    \label{fig:pre_mia}
\end{figure}

To verify the above analysis, we directly apply the objective function of VIB~\cite{alemi2016deep} to GCN and denote the model as \textbf{GCN+IB}. We investigate the performance of GCN+IB against membership inference attacks by varying the number of training labeled nodes. Specifically, we vary the label rates on Cora~\cite{kipf2016semi} by $\{2\%, 4\%, 6\%, 8\%\}$. 
The ROC score of MIA-F~\cite{olatunji2021membership} is used to evaluate the ability to preserve membership privacy. Note that a \textit{lower} MIA-F ROC score indicates better performance in preserving membership privacy. The experimental settings of MIA and the hyperparameter tuning follow the description in Sec.~\ref{sec:exp_set}. The results are presented in Fig.~\ref{fig:pre_mia}, where we can observe that (i) MIA-F ROC of GCN+IB is consistently lower than GCN, which verifies that adopting IB can benefit membership privacy preserving; (ii) membership inference attack can still be very effective on GCN+IB when the label rate is small. With the increase in label rate, the MIA-F ROC score of GCN+IB significantly decreases and the gap between GCN and GCN+IB becomes larger. This empirically shows that abundant labeled samples are required for applying IB to defend MIA effectively.

\subsection{Impacts of IB to Adversarial Robustness} \label{Sec:pre_robust}
Intuitively, the negative effects of adversarial perturbations can be reduced with IB, as IB aims to learn representations that only contain information about the label of the classification task. This has been verified by VIB~\cite{alemi2016deep}, which incorporates IB to deep neural networks on i.i.d data. However, GNNs generally explicitly combine the information of center nodes and their neighbors to obtain node representations. For example, in each layer of GCN, the center node representations are updated by averaging with neighbor representations. Directly using the IB objective function to a GNN encoder may not be sufficient to bottleneck the minimal sufficient neighbor information. As a result, adversarial perturbations on graph structures can still degrade the performance. To empirically verify this, we compare the performance of GCN+IB with GCN on graphs perturbed by Metattack~\cite{zugner2019adversarial} and Nettack~\cite{zugner2018adversarial}. The experimental settings follow the description in Sec.~\ref{sec:exp_set}. The results are shown in Tab.~\ref{tab:pre_robust}. We can observe that the GCN model trained with IB objective function achieves better performance on perturbed graphs, which indicates the potential of giving robust node classification with IB. However, compared with the performance on clean graphs, the accuracy of GCN+IB on perturbed graphs is still relatively poor. This empirically verifies that simply applying IB objective function to the GNN model cannot properly eliminate the noisy information from adversarial edges and there is still a large space to improve IB for robust GNN.
% \enyan{May also add RS-GNN, which have structure denoising mechanism. Poor performance compared with RS-GNN also verify our opinion.}

\begin{table}[t]
    \small
    \centering
    \caption{Results (Accuracy(\%)+std) on perturbed graphs.}
    \vskip -1.5em
    \begin{tabularx}{0.95\linewidth}{p{0.1\linewidth}p{0.1\linewidth}CCC}
    \toprule
    Dataset & Model & Clean & Metattack & Netattack \\
    \midrule
    \multirow{2}{*}{Cora} 
    & GCN    & 73.2 $\pm 0.8$ & 61.9 $\pm 1.4$ & 54.6 $\pm 0.8$ \\
    & GCN+IB & 73.1 $\pm 0.5$ & 66.3 $\pm 0.3$ & 58.0 $\pm 1.6$ \\
    \midrule
    \multirow{2}{*}{Citeseer}
    & GCN    & 72.1 $\pm 0.2$ & 64.1 $\pm 0.5$ & 62.3 $\pm 0.7$ \\
    & GCN+IB & 71.5 $\pm 0.3$ & 66.8 $\pm 1.1$ & 63.1 $\pm 1.3$ \\
    \bottomrule
    \end{tabularx}
    \label{tab:pre_robust}
    \vskip -1em
\end{table}

\section{Methodology}

As analyzed in Sec.~\ref{sec:pre_privacy} and Sec.~\ref{Sec:pre_robust}, information bottleneck can benefit both robustness and membership privacy. However, there are two challenges to be addressed for achieving better robust and membership privacy-preserving predictions: (i) how to design a graph information bottleneck framework that can handle adversarial edges? and (ii) how to ensure membership privacy with IB given a small set of labels? To address these challenges, we propose a novel framework {\method}, which is illustrated in Fig.~\ref{fig:framework}.  
In {\method}, the attribute information and neighbor information are separately bottlenecked. The attribute bottleneck aims to extract  node attribute information relevant to the classification. The neighbor bottleneck aims to control the information flow from neighbors to the center node, and to filter out noisy or useless neighbors for the prediction on the center node. Hence, the influence of adversarial edges can be reduced. 
Moreover, a novel self-supervisor is proposed to guide the training of the neighbor bottleneck to benefit the noisy neighbor elimination. 
To address the challenge of lacking plenty of labels for membership privacy-preserving, we propose to obtain pseudo labels and combine them with provided labels in the training phase. Specifically, {\method} will be trained with the IB objective function with both labels on labeled nodes and pseudo labels on unlabeled nodes to guarantee membership privacy.  More details of the design are presented in the following sections.

\subsection{Graph Information Bottleneck}
\begin{figure}
    \centering
    \includegraphics[width=1\linewidth]{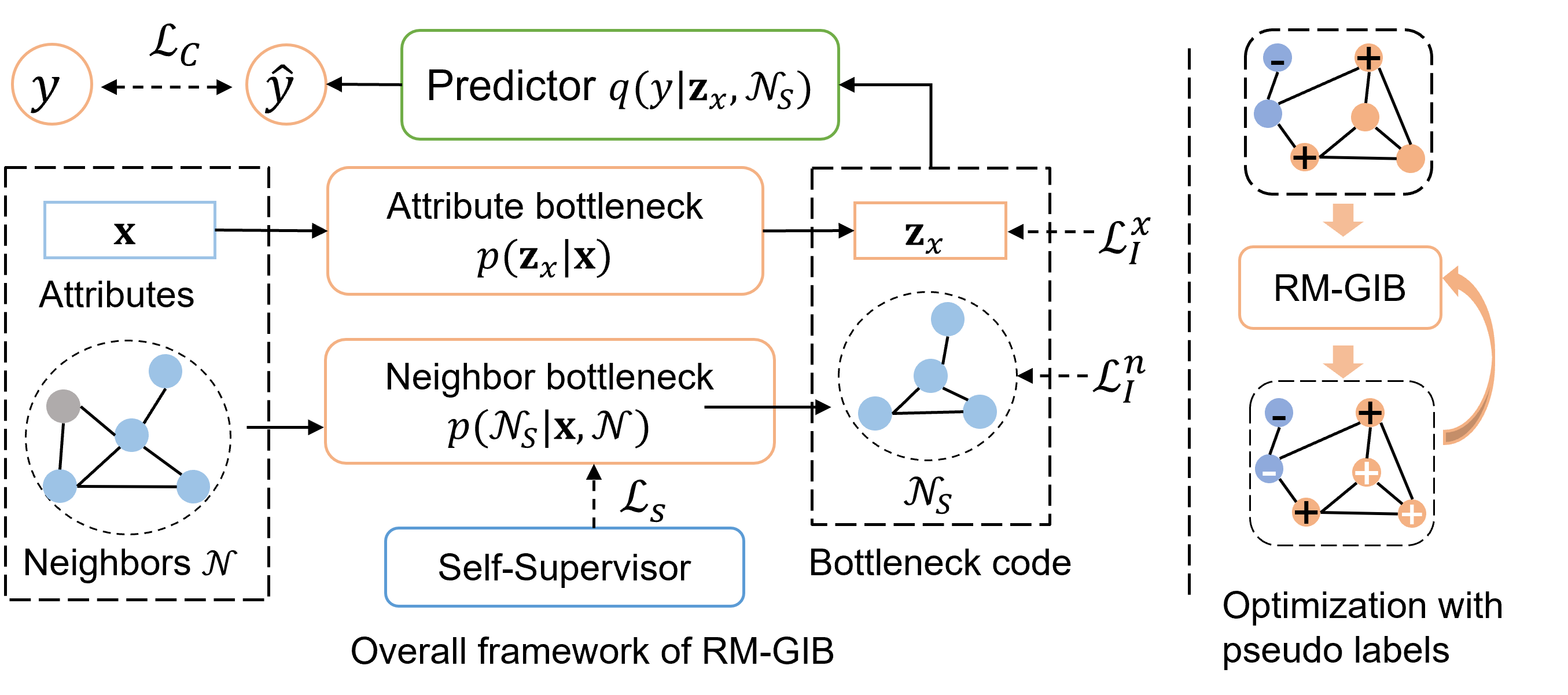}
    \vskip -1em
    \caption{The overall framework of our {method} and the illustration of optimization with pseudo labels.}
    \vskip -1em
    \label{fig:framework}
\end{figure}

In this section, we give the objective of the proposed graph information bottleneck. For graph-structured data, both node attributes and neighbors contain crucial information for node classification. Therefore, for each node $v$, {\method} will extract bottleneck code from both node attributes $\mathbf{x}$ and its neighbor set $\mathcal{N}$, which is shown in Fig.~\ref{fig:framework}. More specifically, the bottleneck code is separated into two parts: (i) $\mathbf{z}_x=f_x(\mathbf{x})$, encoding the node attribute information; (ii) $\mathcal{N}_S=f_n(\mathcal{N},\mathbf{x})$, a subset of $v$'s neighbors that bottleneck the neighborhood information for prediction. Note that $\mathcal{N}$ can be multi-hop neighbors of a node. With the explicit bottleneck mechanisms on both attributes and neighbors, the noisy information from adversarial perturbations can be suppressed. The objective function of the graph information bottleneck is given as: 
\begin{equation}
    \min_{\theta} -I(\mathbf{z}_x, \mathcal{N}_S;y) + \beta I(\mathbf{z}_x, \mathcal{N}_S; \mathbf{x}, \mathcal{N})
    \label{eq:GIB}
\end{equation}
where $\theta$ denotes the learnable parameters of attribute bottleneck and neighbor bottleneck.
However, it is challenging to directly optimize Eq.(\ref{eq:GIB}) due to the difficulty in computing the mutual information. Thus, we derive tractable variational upper bounds of the two terms in Eq.(\ref{eq:GIB}).

Following~\cite{alemi2016deep}, we introduce $q(y|\mathbf{z}_x, \mathcal{N}_S)$ as the parameterized variational approximation of $p(y|\mathbf{z}_x,\mathcal{N}_S)$. Note that $q(y|\mathbf{z}_x, \mathcal{N}_S)$ also can be viewed as a predictor, which can be flexible to various GNNs. Then, the upper bound of $-I(\mathbf{z}_x, \mathcal{N}_S;y)$ can be derived as:
\begin{equation}
    \begin{aligned}
        -I(\mathbf{z}_x, \mathcal{N}_S;y) & \leq \mathbb{E}_{p(\mathbf{z}_x, \mathcal{N}_S,y)}[-\log q(y|\mathbf{z}_x, \mathcal{N}_S)] - H(y) \\
        & \leq \mathbb{E}_{p(\mathbf{z}_x, \mathcal{N}_S,y)}[-\log q(y|\mathbf{z}_x, \mathcal{N}_S)]  = \mathcal{L}_C
    \end{aligned}
    \label{eq:class}
\end{equation}
% where $p_l(\mathbf{z}_x, \mathcal{N}_S,y)$ denotes the distribution on the labeled set $\mathcal{V}_L$. 

Next, we give the upper bound of the second term in Eq.(\ref{eq:GIB}). Since the attribute code $\mathbf{z}_x$ is given by $f_x(\mathbf{x})$  which only takes node attributes as input, we can infer that $p(\mathbf{z}_x|\mathbf{x},\mathcal{N})=p(\mathbf{z}_x|\mathbf{x})$. Then, we can get $p(\mathbf{z}_x, \mathcal{N}_S| \mathbf{x},\mathcal{N})=p(\mathbf{z}_x|\mathbf{x})p(\mathcal{N}_S | \mathbf{x}, \mathcal{N})$, which indicates $I(\mathbf{z}_x, \mathcal{N}_S|\mathbf{x},\mathcal{N})=0$. As a result, $ I(\mathbf{z}_x, \mathcal{N}_S; \mathbf{x},\mathcal{N})$ can be derived to:
\begin{equation} %\small
\begin{aligned}
    & I(\mathbf{z}_x,\mathcal{N}_S;\mathbf{x},\mathcal{N}) = I(\mathbf{z}_x;\mathbf{x},\mathcal{N})+I(\mathcal{N}_S;\mathbf{x},\mathcal{N}|\mathbf{z}_x) \\
   = & I(\mathbf{z}_x; \mathbf{x}) + I(\mathcal{N}_S;\mathbf{x},\mathcal{N}) - I(\mathbf{z}_x;\mathcal{N}_S) + I(\mathbf{z}_x, \mathcal{N}_S|\mathbf{x},\mathcal{N}) \\
   \leq & I(\mathbf{z}_x; \mathbf{x}) + I(\mathcal{N}_S;\mathbf{x},\mathcal{N})
\end{aligned}
\label{eq:info}
\end{equation}
The term $I(\mathbf{z}_x;\mathbf{x})$ in Eq.(\ref{eq:info}) can be upper bounded as:
\begin{equation}
    I(\mathbf{z}_x; \mathbf{x}) \leq \mathbb{E}_{p(\mathbf{x})}[KL(p(\mathbf{z}_x|\mathbf{x})||q(\mathbf{z}_x))]  = \mathcal{L}_I^x
\end{equation}
where $q(\mathbf{z}_x)$ is the variational approximation to the marginal $p(\mathbf{z}_x)$ $KL$ denotes the KL divergence.  $q(\mathbf{z}_x)$ is flexible to various distributions such as normal distribution. 
Similarly, let $q(\mathcal{N}_S)$ be the variational approximation to the marginal $p(\mathcal{N}_S)$, the upper bound of $I(\mathcal{N}_S;\mathbf{x},\mathcal{N})$ is given as:
\begin{equation}
    I(\mathcal{N}_S; \mathbf{x},\mathcal{N}) \leq \mathbb{E}_{p(\mathbf{x}, \mathcal{N})}[KL(p(\mathcal{N}_S|\mathbf{x},\mathcal{N})||q(\mathcal{N}_S))] =  \mathcal{L}_I^n
\end{equation}
With the above derivations, we obtain a variational upper bound of  Eq.(\ref{eq:GIB}) as the objective function of graph information bottleneck:
\begin{equation} 
    \min_{\theta} \mathcal{L}_C + \beta (\mathcal{L}_I^x + \mathcal{L}_I^n)
    \label{eq:GIB_bound}
\end{equation}
where $\theta$ denotes the parameters to be optimized in the graph information bottleneck. 
\subsection{Neural Network Parameterization}
With the objective function of graph information bottleneck given above, we specify the neural network parameterization of the attribute bottleneck $p(\mathbf{z}_x|\mathbf{x})$, neighbor bottleneck $p(\mathcal{N}_S|\mathbf{x},\mathcal{N})$ and the predictor $q(y|\mathbf{z}_x,\mathcal{N}_S)$ in this subsection.

\subsubsection{Attribute Bottleneck}
The attribute bottleneck aims to learn a code $\mathbf{z}_x$ that contains minimal and sufficient information for classification from node attributes $\mathbf{x}$. Inspired by~\cite{alemi2016deep},  a MLP model and reparameterization trick is adopted to model $p(\mathbf{z}_x|\mathbf{x})$ for attribute bottleneck. Specifically, we assume $p(\mathbf{z}_x|\mathbf{x})$ follows Gaussian distribution with the mean and variance as the output of a MLP:
\begin{equation}
    p(\mathbf{z}_x|\mathbf{x})=N(\mathbf{z}_x; \boldsymbol{\mu}, \boldsymbol{\sigma}^2 \mathbf{I}), \quad \boldsymbol{\mu}, \boldsymbol{\sigma} = f_x(\mathbf{x})
\end{equation}
where $f_x$ is a MLP which outputs $\boldsymbol{\mu}$ and $\boldsymbol{\sigma}$ as the mean and standard deviation. $\mathbf{z}_x$ can be sampled by $\mathbf{z}_x = \boldsymbol{\mu} + \boldsymbol{\sigma} \odot \boldsymbol{\epsilon}$, where $\boldsymbol{\epsilon}$ is sampled from the normal distribution $N(\mathbf{0},\mathbf{I})$. As $q(\mathbf{z}_x)$ is set as normal distribution, $KL(p(\mathbf{z}_x|\mathbf{x})||q(\mathbf{z}_x))$ can be easily computed for $\mathcal{L}_I^x$.

\subsubsection{Neighbor Bottleneck} For the neighbor bottleneck, it will extract a subset of neighbors that are useful for the target classification task. With an ideal neighbor bottleneck, noisy neighbors caused by adversarial edges and inherent structural noise can be eliminated. Here, we propose a parameterized neighbor bottleneck to model $p(\mathcal{N}_S|\mathbf{x},\mathcal{N})$. To ease the difficulty of computation, we decompose $p(\mathcal{N}_S|\mathbf{x},\mathcal{N})$ into a multivariate Bernoulli distribution as 
\begin{equation}
    p(\mathcal{N}_S|\mathbf{x},\mathcal{N})={\prod}_{u \in \mathcal{N}_S} p_u {\prod}_{u \in \mathcal{N}\backslash \mathcal{N}_S} (1-p_u)
\end{equation}
%$p(\mathcal{N}_S|\mathbf{x},\mathcal{N})=\prod_{u \in \mathcal{N}} p_u$, 
where $p_u$ is the probability of $p(u|\mathbf{x},\mathcal{N})$ that follows Bernoulli distribution.
To ensure the gradients can be propagated from the classifier to the neighbor bottleneck module during the optimization, Gumbel-Softmax trick~\cite{jang2016categorical} with the temperature set as 1 is applied in the sampling phase.
Each $p_u$ will be estimated by a MLP which takes the center node attributes $\mathbf{x}$ and the attributes of the neighbor $\mathbf{x}_u$ as input by:
\begin{equation}
    p_u = \sigma(\mathbf{h}_u^T \mathbf{h}) \text{ with } ~~ \mathbf{h} = f_n(\mathbf{x}),~~ \mathbf{h_u} = f_n(\mathbf{x}_u),
    \label{eq:pu}
\end{equation}
where $\sigma$ denotes the sigmoid  function, and $f_n$ denotes a MLP model. As for the variational approximation of marginal distribution $q(\mathcal{N}_S)$, we also use a multivariate Bernoulli distribution $q(\mathcal{N}_S)=r^{|\mathcal{N}_S|}(1-r)^{|\mathcal{N}|-|\mathcal{N}_S|}$  where $r \in [0, 1]$ is the probability of a predefined Bernoulli distribution. 
Then, the information loss on neighbor bottleneck $\mathcal{L}_I^n$ in Eq.(\ref{eq:GIB_bound}) can be computed as:
\begin{equation}
    \mathcal{L}_I^n  = \mathbb{E}_{p(\mathbf{x}, \mathcal{N})}\big[\sum_{u\in \mathcal{N}} p_u \log \frac{p_u}{r} + (1-p_u) \log \frac{1-p_u}{1-r}\big].
\end{equation}

\subsubsection{Predictor} The predictor $q(y|\mathbf{z}_x, \mathcal{N}_S)$ will give predictions based on the bottleneck code of attributes and the extracted subset of neighbors. To fully utilize the rich information from bottlenecked neighbors, a GNN model is deployed as the predictor in {\method}. It is flexible to adopt various GNN models such as GCN~\cite{kipf2016semi} and SGC~\cite{wu2019simplifying}. Note that if $\mathcal{N}_S$ contains neighbors in $K$ hops, a $K$ layer GNN will be adopted in this situation. In addition, to avoid the influence of noises in attributes, we also use the attribute bottleneck code $\mathbf{z}_u$ for each neighbor $\mathbf{z}_u \in \mathcal{N}_S$. Let $\mathbf{A}_S$ denote the local adjacency matrix that connects nodes in $\mathcal{N}_S$ and the center node, the prediction can be formally defined as:
\begin{equation}
    \hat{y} = f_c(\mathbf{z}_x, \{\mathbf{z}_u\}_{u \in \mathcal{N}_S}, \mathbf{A}_S),
\end{equation}
where $f_c$ is the GNN-based classifier. As the prediction is given on bottlenecked attributes and neighbors, it can give robust predictions against adversarial perturbations on attributes and graph structures.
% Given a labeled set $\mathcal{V}_L$, the $\mathcal{L}_C$ in Eq.(\ref{eq:class}) can be rewritten as:
% \begin{equation}
%     \mathcal{L}_C = \frac{1}{|\mathcal{V}_L|}\sum_{v \in \mathcal{V}_L} l(y_v, \hat{y}_v),
% \end{equation}
% where $l(\cdot)$ denotes the cross entropy loss and $\hat{y}_v$ denotes the predictions on node $v$ with $f_c$.

\subsection{Self-supervision for Neighbor Bottleneck}
The objective function in Eq.(\ref{eq:GIB_bound}) will force the neighbor bottleneck to extract minimal sufficient neighbors that achieve good classification performance. However, the training of neighbor bottleneck will only rely on the implicit supervision from the small set of labels in semi-supervised node classification, which may not be sufficient to train a neighbor bottleneck to handle various structural noises. Therefore, we propose a novel self-supervisor to explicitly guide the training of the neighbor bottleneck. The major intuition is that the neighbor nodes with low mutual information with the center node are likely to be the noisy neighbors that are not helpful for the prediction on the center nodes. Hence, we can first estimate the mutual information of each pair of linked nodes. Then, neighbors with low mutual information scores with the center node can be viewed as negative samples and others as positive samples. Next, we give the details of the mutual information estimation followed by the self-supervision loss on the neighbor bottleneck. 

Following~\cite{hjelm2018learning}, a neural network $f_M$ is used to estimate the mutual information between node $v$ and $u$ by:
\begin{equation}
    s_{vu} = \sigma({\mathbf{h}_v^m}^T \mathbf{h}_u^m),~~ \mathbf{h}_v^m = f_M(\mathbf{x}_v), ~~ \mathbf{h}_u^m = f_M(\mathbf{x}_u),
\end{equation}
where $\sigma$ is the sigmoid activation function and $f_M$ is an MLP instead of a GNN model to avoid the negative effects of inherent and adversarial structural noises. A larger $s_{vu}$ indicates higher point-wise mutual information between $v$ and $u$. The mutual information estimator $f_M$ can be trained with the following objective~\cite{hjelm2018learning}:
\begin{equation}
    \min_{\theta_M} -\frac{1}{|\mathcal{V}|} \sum_{v \in \mathcal{V}} \sum_{u \in \mathcal{N}_v}[-\log(s_{vu})-\mathbb{E}_{n \sim p(v)} \log (1-s_{vn})],
    \label{eq:muta}
\end{equation}
where $\theta_M$ represents parameters of $f_M$ and $\mathcal{N}_v$ is the set of neighbors of $v$. $p(v)$ is the distribution of sampling negative samples for $v$, which is set as a uniform distribution. With Eq.(\ref{eq:muta}), the mutual information estimator can be trained. Then, we can select the neighbors with a mutual information score lower than the threshold as the negative pairs for neighbor bottleneck. Specifically, for each node $v$, the negative neighbors can be obtained by:
\begin{equation}
    \mathcal{N}_v^- = \{u \in \mathcal{N}_v; s_{vu} < T\},
\end{equation}
where $T$ is the predefined threshold. With the negative neighbors, the self-supervision on neighbor bottleneck can be given by:
\begin{equation}
    \min_{\theta} \mathcal{L}_S = \frac{1}{|\mathcal{V}|} \sum_{v \in \mathcal{V}} \big[ \sum_{u \in \mathcal{N}_v^+}-\log(p_u^v) - \sum_{u \in \mathcal{N}_v^-}\log(1-p_u^v)\big],
    \label{eq:self}
\end{equation}
where $\theta$ denotes parameters of {\method}, $\mathcal{N}_v^+ = \mathcal{N}_v-\mathcal{N}_v^-$ and $p_u^v$ corresponds to the probability value of $p(u|\mathbf{x}_v, \mathcal{N}_v)$ given by neighbor bottleneck thorough Eq.(\ref{eq:pu}). 
With Eq.(\ref{eq:self}), the neighbors who are likely to be noisy will be given lower probability scores in the neighbor bottleneck. 

\subsection{Privacy-Preserving Optimization with Pseudo Labels}
As empirically verified in Sec.~\ref{sec:pre_privacy}, a large number of labels are required to preserve membership privacy with IB. Thus, we propose to obtain pseudo labels of unlabeled nodes to enlarge the training set to further improve membership privacy. In particular, the adoption of pseudo labels in {\method} can benefit the membership privacy in two aspects: (i) classification loss will also be optimized with unlabeled nodes, which increases the confidence scores of prediction on unlabeled nodes. This will make it more difficult to distinguish the prediction vectors of labeled and unlabeled nodes. (ii) involving a large number of unlabeled nodes in the training can improve the generalization ability of attribute and neighbor bottleneck, which can help narrow the gap between the predictions on training samples and test samples. Moreover, the improvement of bottleneck code can also benefit the classification performance. Next, we give the details of the pseudo label collection and the optimization with pseudo labels.

To obtain pseudo labels that are robust to noises in graphs, we can train {\method} with the IB objective function combined with the self-supervision on neighbor bottleneck. Let $\mathcal{L}_C(\mathcal{V}_L, \mathcal{Y}_L)$, $\mathcal{L}_I^x(\mathcal{V}_L)$, and $\mathcal{L}_I^n(\mathcal{V}_L)$ denote the three terms in the IB objective function in Eq.(\ref{eq:GIB_bound}) on the labeled set $\mathcal{V}_L$. Then, the process of training {\method} for pseudo label collection can be formulated as:
 \begin{equation}
     \min_{\theta} \mathcal{L}_C(\mathcal{V}_L, \mathcal{Y}_L) + \beta\big(\mathcal{L}_I^x(\mathcal{V}_L) + \mathcal{L}_I^n(\mathcal{V}_L)\big) + \gamma \mathcal{L}_S,
     \label{eq:pseudo}
 \end{equation}
where $\beta$ and $\gamma$ are hyperparameters to control the contributions of regularization on bottleneck code and the self-supervision on neighbor bottleneck. $\theta$ denotes the learnable parameters in {\method}.
With the {\method} trained on Eq.(\ref{eq:pseudo}), we can collect high-quality pseudo labels $\mathcal{\hat Y}_U$ of the unlabeled set $\mathcal{V}_U$. Then, we combine pseudo labels $\mathcal{\hat Y}_U$ with provided labels $\mathcal{Y}_L$ and retrain {\method} for membership privacy-preserving.
Let $\mathcal{V}_P = \mathcal{V}_L \cup \mathcal{V}_U$ and $\mathcal{\hat Y}_P=\mathcal{\hat Y}_U \cup \mathcal{Y}_L$ denote the enlarged labeled node set and labels,  the membership privacy-preserving optimization can be formally written as:
 \begin{equation}
     \min_{\theta} \mathcal{L}_C(\mathcal{V}_P, \mathcal{\hat Y}_P) + \beta(\mathcal{L}_I^x(\mathcal{V}_P) + \mathcal{L}_I^n(\mathcal{V}_P)) + \gamma \mathcal{L}_S
     \label{eq:final}
 \end{equation}
The hyperparameters $\beta$ and $\gamma$ are set the same as Eq.(\ref{eq:pseudo}). 

\begin{table}[t]
    \small
    \caption{Statistics of datasets.}
    \vskip-1.5em
    \centering
    \begin{tabularx}{0.8\linewidth}{p{0.15\linewidth}CCCC}
    \toprule
         & Cora &  Citeseer & Pubmed & Flickr  \\
    \midrule
    \#classes & 7 & 6 & 3 & 7\\
    \#features & 1,433 & 3,703 & 500 & 500\\
    \#nodes & 2,485  & 2,110 & 19,717 & 89,250\\
    \#edges & 5,069  & 3,668 & 44,338 & 899,756\\
    \bottomrule
    \end{tabularx}
    \label{tab:dataset}
    \vskip -1em
\end{table}
\begin{table*}[t]
    \small
    \centering
    \caption{Comparison with baselines in defending membership inference attack on various clean graphs.}
    \vskip -1 em
    \begin{tabularx}{0.985\textwidth}{p{0.05\textwidth}p{0.12\textwidth}CCCCCCCC}
    \toprule
    Dataset & Metrics & GCN & GCN+PL & Adv-Reg & DP-SGD & GIB &LBP & NSD & {\method} \\
    \midrule
    \multirow{3}{*}{Cora}
        & Accuracy (\%) $\uparrow$    & 73.2$\pm 0.8$ & {74.7$\pm 0.2$} & \underline{75.5 $\pm 0.8$} & 57.9 $\pm 0.2$	& 72.5 $\pm 0.7$ & 69.7 $\pm 0.7$ &	65.4 $\pm 0.3$ & \textbf{78.1} $\bf \pm 0.4$ \\
        & MIA-F ROC (\%) $\downarrow$ & 90.6 $\pm 0.8$ &	\underline{61.6$\pm 0.2$} &	70.6 $\pm 0.4$ & 73.8 $\pm 3.3$ & 86.6 $\pm 0.8$ & 71.0 $\pm 1.7$ & 81.8 $\pm 0.8$ & \textbf{57.4} $\bf \pm 0.2$ \\
        & MIA-S ROC  (\%) $\downarrow$ & 88.8 $\pm 0.2$ &	\underline{63.8 $\pm 0.8$} &	70.6 $\pm 0.3$ & 75.3 $\pm 1.2$ &	87.3 $\pm 0.7$ & 71.1 $\pm 1.5$ &	81.2 $\pm 0.6$	& \textbf{59.5} $\bf \pm 1.2$
 \\
    \midrule
    \multirow{3}{*}{Citeseer}
        & Accuracy (\%)$\uparrow$    & 72.1 $\pm 0.2$ &	\underline{73.1 $\pm 0.2$} & {72.4 $\pm 1.0$} & 57.9 $\pm 0.2$ & 71.0 $\pm 0.2$  & 66.5 $\pm 0.8$ & 65.6 $\pm 0.2$ &  \textbf{73.9} $\bf \pm 0.6$  \\
        & MIA-F ROC (\%)$\downarrow$ & 88.5 $\pm 1.8$ & 65.2 $\pm 0.6$ & \underline{60.9 $\pm 0.6$} & 73.8 $\pm 3.3$ & 85.8 $\pm 0.5$  & 66.6 $\pm 0.4$ & 84.4 $\pm 0.1$ &  \textbf{55.2} $\bf \pm 0.8$  \\
        & MIA-S ROC (\%)$\downarrow$ & 84.9 $\pm 1.5$ & 65.8 $\pm 0.5$ & \underline{61.2 $\pm 1.1$} & 75.3 $\pm 1.2$ & 80.3 $\pm 0.4$	& 67.3 $\pm 0.7$ & 88.3 $\pm 0.1$  & \textbf{55.9} $\bf \pm 1.7$  \\
    \midrule
    \multirow{3}{*}{Pubmed}
        & Accuracy (\%)$\uparrow$    & \underline{79.9 $\pm 0.1$} & 79.9 $\pm 0.1$ & 79.4 $\pm 1.1$ & 69.3 $\pm 3.2$ & 78.1 $\pm 0.4$ & 78.3 $\pm 0.1$ & 75.5 $\pm 0.1$ & \textbf{81.4} $\bf \pm 0.2$ \\
        & MIA-F ROC (\%)$\downarrow$ & 75.1 $\pm 0.2$ &	60.8 $\pm 0.2$ & 60.6 $\pm 1.8$ & \underline{56.3 $\pm 1.8$} & 68.5 $\pm 1.6$ &	67.4 $\pm 1.6$ & 68.4 $\pm 0.2$ & \textbf{53.9}  $\bf \pm 0.3$\\
        & MIA-S ROC (\%)$\downarrow$ & 73.4 $\pm 0.1$ & 63.4 $\pm 0.2$ & 62.8 $\pm 2.0$ & \underline{58.3 $\pm 2.1$} & 67.0 $\pm 1.8$ & 65.7 $\pm 2.0$ & 72.1 $\pm 0.1$ & \textbf{57.2} $\bf \pm 0.2$ \\
    \midrule
    \multirow{3}{*}{Flickr}
        & Accuracy (\%)$\uparrow$    & \textbf{52.5} $\bf \pm 0.2$ & 51.8 $\pm 0.8$ & 48.2 $\pm 1.8$ & 46.2 $\pm 0.1$ & 45.2 $\pm 2.0$ & 44.6 $\pm 0.5$ & 41.6 $\pm 0.5$ & \underline{52.2 $ \pm 0.2$}\\
        & MIA-F ROC (\%)$\downarrow$ & 87.9 $\pm 0.7$ & 72.9 $\pm 1.5$ & 64.3 $\pm 3.9$ & 66.5 $\pm 0.7$ & 79.9 $\pm 4.4$ & 67.9 $\pm 0.8$ & \underline{59.0 $\pm 1.5$} & \textbf{58.2} $\bf \pm 0.1$\\
        & MIA-S ROC (\%)$\downarrow$ & 84.2 $\pm 0.7$ & 69.7 $\pm 1.2$ & 66.4 $\pm 1.2$ & 65.1 $\pm 0.6$ & 76.5 $\pm 0.7$ & 71.3 $\pm 0.9$ & \underline{63.5 $\pm 1.3$} & \textbf{57.6} $\bf \pm 0.3$\\
    \bottomrule
    \end{tabularx}
    
    \label{tab:results_mia}
    \vskip -1em
\end{table*}

\section{Experiments}
In this subsection, we evaluate the proposed {\method} on various real-world datasets to answer the following research questions:
\begin{itemize}[leftmargin=*]
    \item \textbf{RQ1} Can our proposed {\method} preserve the membership privacy in node classification given a small set of labeled nodes?
    \item \textbf{RQ2} Is {\method} robust to  adversarial perturbations on graphs and can membership privacy be simultaneously guaranteed?
    \item \textbf{RQ3} How does  each component of {\method} contribute to the robustness and membership privacy?
\end{itemize} 
\subsection{Experimental Settings}
\label{sec:exp_set}

\subsubsection{Datasets}
We conduct experiments on widely used publicly available benchmark datasets, i.e., Cora, Citeseer, Pubmed~\cite{kipf2016semi}, and Flickr~\cite{zeng2020graphsaint}. The key statistics of these datasets can be found in Tab.~\ref{tab:dataset}. Details of the dataset settings can be found in Appendix~\ref{app:dataset}

% Cora, Citeseer, and Pubmed are citation networks, where nodes in the graphs represent the papers and edges denote citation relationship. The attributes of the nodes are the bag-of-words of these papers. For small citation graphs, i.e., Cora and Citeseer, we randomly sample 2\% nodes as the training set. For the large citation graph Pubmed, we randomly sample 0.5\% nodes as the training set.  As for Flickr~\cite{zeng2020graphsaint}, it is a large-scale graph to categorize the type of images. Each node represents an image and the image description is used as a node attribute. Edges are formed between nodes sharing common properties. We randomly sample 2\% nodes from Flickr as the training set. Splits of validation and test sets on all datasets follow the cited papers for consistency. Note that the training node set doesn't overlap with the validation and test sets.

\subsubsection{Baselines} To evaluate the performance in preserving membership privacy, we compare {\method} with the  representative graph neural network \textbf{GCN}~\cite{kipf2016semi} and an existing work of graph information bottleneck \textbf{GIB}~\cite{wu2020graph}. We also incorporate a state-of-the-art regularization method, i.e., adversarial regularization~\cite{nasr2018machine} (\textbf{Adv-Reg}). A differential privacy-based method \textbf{DP-SGD}~\cite{abadi2016deep} is also compared. Additionally, we compare two recent methods for defending membership inference attacks on GNNs, which are \textbf{LBP}~\cite{olatunji2021membership} and \textbf{NSD}~\cite{olatunji2021membership}. LBP adds noise to the posterior before it is released to end users. NSD randomly chooses neighbors of the queried node to limit the amount of information used in the target model for membership privacy protection.

To evaluate the robustness of {\method} against adversarial attacks on graphs, apart from GCN and GIB, we also compare representative and state-of-the-art robust GNNs. Specifically, we compare two classical preprocessing methods, i.e., \textbf{GCN-jaccard}~\cite{wu2019adversarial} and \textbf{GCN-SVD}~\cite{entezari2020all}. Two state-of-the-art robust GNNs are also incorporated in the comparison, which are \textbf{Elastic}~\cite{liu2021elastic} and \textbf{RSGNN}~\cite{dai2022towards}. For more detailed descriptions about the above baselines, please refer to Appendix~\ref{app:baseline}. To make a fair comparison, the hyperparameters of all baselines are tuned based on the validation set. For our {\method}, hyperparameter sensitivity analysis is given in Sec.~\ref{sec:hyper}. More implementation details of {\method} can be found in Appendix~\ref{app:imple}.

% \begin{table}[t]
%     \small
%     \caption{Accuracy (\%) of nodes not in the training graph.}
%     \vskip-1em
%     \centering
%     \begin{tabularx}{0.95\linewidth}{XCCCC}
%     \toprule
%          & Cora &  Citeseer & Pubmed & Flickr  \\
%     \midrule
%     GCN     & 72.1 $\pm 0.4$ & 71.9 $\pm 1.5$ & \textbf{79.4} $\pm 0.1$ & 52.0 $\pm 2.4$ \\
%     GIB     & 71.2 $\pm 0.2$ & 70.0 $\pm 2.1$ & 77.9 $\pm 0.5$ & 48.4 $\pm 2.2$ \\
%     Adv-Reg &  74.1 $\pm 0.8$ & 72.0 $\pm 1.2$ &78.2 $\pm 0.9$ & 47.5 $\pm 0.1$ \\
%     {\method} & \textbf{76.8} $\bf \pm 1.1$ & \textbf{72.0} $\bf \pm 0.2$ & 79.0 $\bf \pm 0.2$ & \textbf{52.1} $\bf \pm 0.1$\\
%     \bottomrule
%     \end{tabularx}
%     \label{tab:inductive}
%     % \vskip -1em
% \end{table}

\subsubsection{Evaluation Protocol} \label{sec:evaluation}
In this subsection, we provide details of experimental settings and metrics to evaluate the performance in defending membership inference attacks and adversarial attacks.

\vspace{0.2em}
\noindent \textbf{Membership Privacy.}
We adopt the state-of-the-art MIA on GNNs in~\cite{olatunji2021membership} for membership privacy-preserving evaluation. The shadow training~~\cite{olatunji2021membership} described in Sec.~\ref{sec:mia} is adopted. Here, GCN is applied as the shadow model. The attack setting is set as black-box, i.e., the attacker can only obtain the predictive vectors and cannot access model parameters. As for the shadow dataset, we use two settings: 
\begin{itemize}[leftmargin=*]
    \item \textbf{MIA-F}: The attacker has the complete graph used for training along with a small set of labels; 
    \item \textbf{MIA-S}: The attacker has a subgraph of the dataset with a small set of labels; In all experiments, we randomly sample 50\% nodes as the subgraph that is available for the attacker.
\end{itemize}
In both settings, \textit{the labeled nodes used in the attack have no overlap with the training set of target model}. The number of labeled nodes used in the attack is the same as the training set.
The attack ROC score is used as a metric for membership privacy-preserving evaluation. And a GNN model with a lower attack ROC score indicates better performance in defending MIAs. 

\vspace{0.2em}
\noindent \textbf{Robustness.} To evaluate the robustness against adversarial attacks, we evaluate {\method} on graphs perturbed by following methods:
\begin{itemize} [leftmargin=*]
    \item \textbf{Mettack}~\cite{zugner2019adversarial}: It aims to reduce the overall performance of the target GNN by perturbing attributes and graph structures. The perturbation rate is set as 0.2 in all experiments.
    \item \textbf{Nettack}~\cite{zugner2018adversarial}: It aims to lead the GNN to misclassify target nodes. Following~\cite{dai2022towards}, 15\% nodes are randomly selected as target nodes. 
\end{itemize}
As the cited papers do, both Mettack and Nettack can access the whole graph. Similar to MIA, the adversarial attacker is assumed to have nodes with labels that do not overlap with the training set.

\subsection{Privacy Preserving on Clean Graphs} \label{sec:exp_mia}
To answer \textbf{RQ1}, we compare {\method} with baselines in defending membership inference attacks on various real-world graphs.
The prediction accuracy of each method is reported. As described in Sec.~\ref{sec:evaluation}, for membership privacy-preserving evaluation, we report the membership attack ROC score on two different settings, i.e., MIA-F and MIA-S, which correspond to the MIA-F ROC and MIA-S ROC in the evaluation metrics. 
Note that lower attack ROC score indicates better performance in preserving privacy. The results on the default dataset split setting described in Appendix~\ref{app:dataset} are reported in Tab.~\ref{tab:results_mia}. Results on different sizes of training set can be found in Appendix~\ref{app:add_label}. From the Tab.~\ref{tab:results_mia}, we can observe:
\begin{itemize}[leftmargin=*]
    \item GCN can be easily attacked by membership inference attacks. This demonstrates the necessity of developing membership privacy-preserving methods for node classification on graphs.
    \item {\method} gives significantly lower scores in MIA-F ROC and MIA-S ROC than baselines. The attack ROC scores can be even  close to 0.5, indicating invalid privacy attacks. This demonstrates the effectiveness of {\method} in preserving membership privacy. 
    % \item Compared with GIB, {\method} perform much better in both accuracy and membership privacy. This is because pseudo labels are incorporated in {\method}, which can helps to bottleneck redundant information to improve performance and narrow the gap between training and test samples for membership privacy.  
    \item The baseline methods often improve membership privacy with a significant decline in accuracy. By contrast, our {\method} can simultaneously maintain high prediction accuracy and preserve membership privacy. This is because baselines generally need to either largely regularize the model or inject strong noises. {\method} does not only rely on the regularization in the IB objective function. Pseudo labels are further incorporated in training {\method}, which helps to bottleneck redundant information to improve performance and narrow the gap between training and test samples for preserving membership privacy.
    % \enyan{to add} 
\end{itemize}
% •	The variant of our method without using pseudo labeling show significant better results in privacy attack defense. This verifies the effectiveness of information bottleneck in membership privacy preserving.
% •	Our method achieve much better results in defending MIA than the variant without using pseudo labeling. This proves the effectiveness of incorporating pseudo labels for membership privacy-preserving.

% \enyan{to add section in inductive setting}
% Membership inference attacks generally aim to attack the GNN model that provides prediction service online. In the online service scenario, it is likely that users query new samples that are not covered in $\mathcal{G}$. Hence, we also evaluate the prediction accuracy of {\method} in an inductive setting, where 10\% nodes are masked out in the training. Then, we report the accuracy on nodes that are masked. The results are shown in Tab.~\ref{tab:inductive}. We can observe that {\method} can also outperform baselines in the inductive setting,  which is consistent with the observations in Tab.~\ref{tab:results_mia}.  

\begin{table*}[t]
    \small
    \centering
    \caption{Comparison with Robust GNNs in node classification (Accuracy(\%)$\pm$Std) on various adversarially perturbed graphs.}
    \vskip -1em
    \begin{tabularx}{0.985\textwidth}{p{0.05\textwidth}p{0.08\textwidth}CCCCCCC}
    \toprule
    Dataset & Graph & GCN & GIB & GCN-jaccard & GCN-SVD & Elastic & RSGNN & {\method} \\
    \midrule
    
    \multirow{3}{*}{Cora}
        & Clean     &73.2 $\pm 0.8$ & 72.5 $\pm 0.7$ & 68.9 $\pm 0.6$ &	65.1 $\pm 0.6$ & \underline{77.9 $\pm 0.9$} & 74.6 $\pm 1.0$ & \textbf{78.5} $\bf \pm 0.6$ \\
        & Metattack & 61.9 $\pm 1.4$ & 65.6 $\pm 0.1$ &	64.4 $\pm 0.2$ & 60.5 $\pm 1.3$ & \underline{70.2 $\pm 0.4$} &	65.3 $\pm 2.5$ & \textbf{71.1} $\bf \pm 0.6$ \\
        & Nettack   & 54.6 $\pm 0.8$ & 60.1 $\pm 3.2$ & 58.6 $\pm 0.5$ & 54.8 $\pm 0.7$ &	64.8 $\pm 1.1$ & \textbf{66.9} $\bf \pm 0.4$ & \underline{65.6 $\pm 1.3$} \\
    \midrule
    \multirow{3}{*}{Citeseer}
        & Clean     & 72.1 $\pm 0.2$ & 71.0 $\pm 0.2$  & 72.2 $\pm 0.1$ & 63.0 $\pm 0.4$ &	\underline{73.7 $\pm 0.3$} & {73.7 $\pm 1.3$} & \textbf{73.9} $\bf \pm 0.6$ \\
        & Metattack & 64.1 $\pm 0.5$ & 66.8 $\pm 0.7$  & 70.5 $\pm 0.1$ & 59.7 $\pm 1.1$ & 71.5 $\pm 0.4$ & \textbf{73.0} $\bf \pm 0.3$ & \underline{72.1 $\pm 0.9$} \\
        & Nettack   & 62.3 $\pm 0.7$ & 63.8 $\pm 1.6$  & 68.9 $\pm 0.2$ & 55.6 $\pm 1.1$ & 68.5 $\pm 0.2$ & \underline{69.0 $\pm 0.9$} & \textbf{69.9} $\bf \pm 0.8 $  \\
    \midrule
    \multirow{3}{*}{Pubmed}
        & Clean     & 79.8 $\pm 0.1$ & 78.1 $\pm 0.4$	& 79.5 $\pm 0.1$ & 75.1 $\pm 0.1$ &	\underline{80.6 $\pm 0.2$} & 75.6 $\pm 0.3$ & \textbf{81.4} $\bf \pm 0.1$ \\
        & Metattack & 67.5 $\pm 0.1$ & 61.5 $\pm 0.4$	& 74.1 $\pm 0.6$ & 74.5 $\pm 0.1$ & 73.5 $\pm 0.2$ & \underline{74.4 $\pm 0.2$} & \textbf{77.3} $\bf \pm 0.1$ \\
        & Nettack   & 68.2 $\pm 0.1$ & 67.5 $\pm 0.6$  & \underline{74.0 $\pm 0.7$} & 67.9 $\pm 0.2$ & 73.2 $\pm 0.3$ & 72.8 $\pm 0.6$ & \textbf{75.0} $\bf \pm 0.2$ \\
    \bottomrule
    \end{tabularx}
    \label{tab:results_robust}
    \vskip -1em
\end{table*}

\begin{figure}[t]
    \small
    \centering
    % \begin{subfigure}{0.49\linewidth}
    %     \includegraphics[width=1\linewidth]{figure/Cora_atk_acc.pdf}
    %     \vskip -1em
    %     \caption{Accuracy on Cora}
    % \end{subfigure}
    \begin{subfigure}{0.49\linewidth}
        \includegraphics[width=0.96\linewidth]{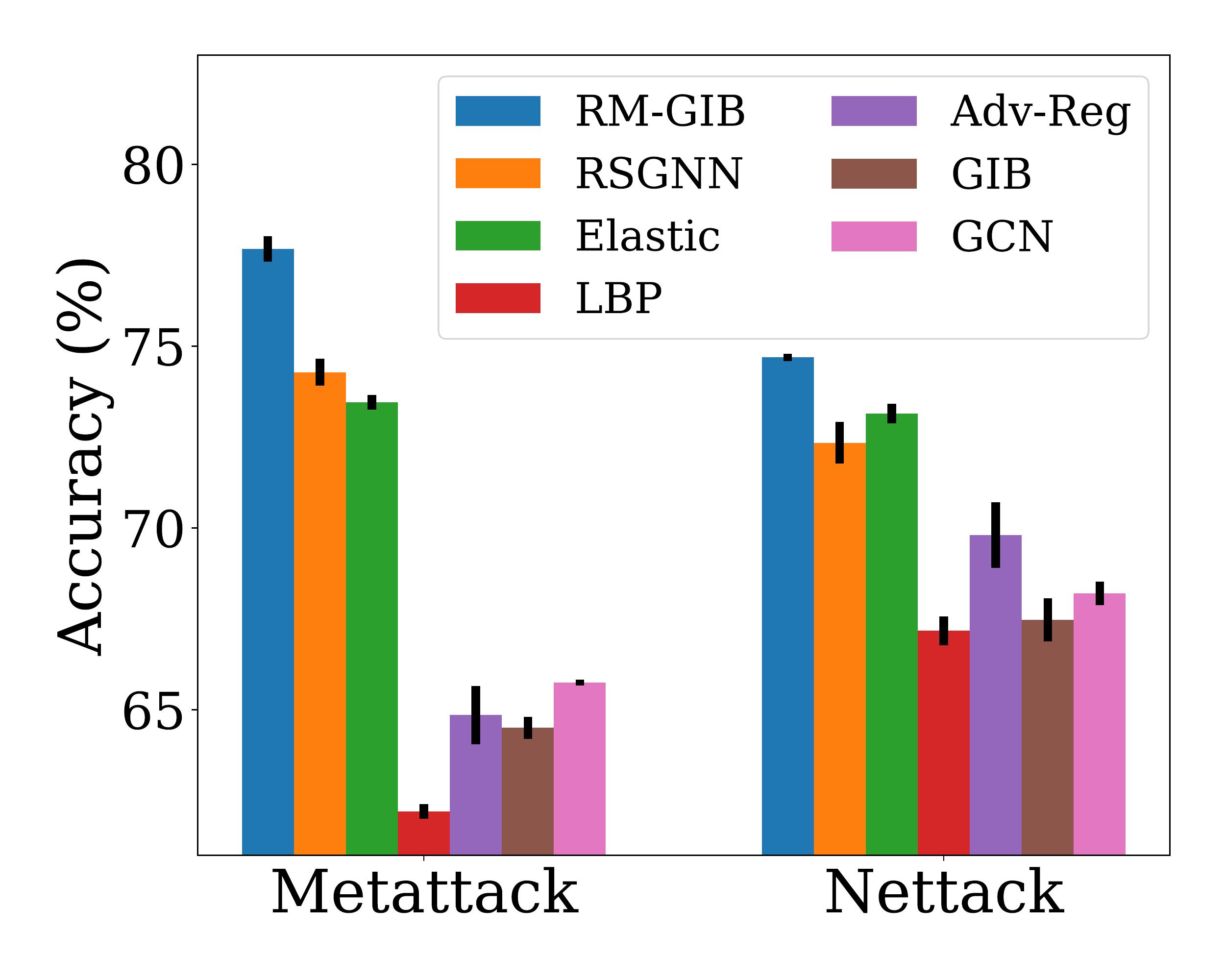}
        \vskip -1em
        \caption{Accuracy on Pubmed}
    \end{subfigure}
    % \begin{subfigure}{0.49\linewidth}
    %     \includegraphics[width=0.96\linewidth]{figure/Cora_atk_mia.pdf}
    %     \vskip -1em
    %     \caption{MIA-F ROC on Cora}
    % \end{subfigure}
    \begin{subfigure}{0.49\linewidth}
        \includegraphics[width=0.96\linewidth]{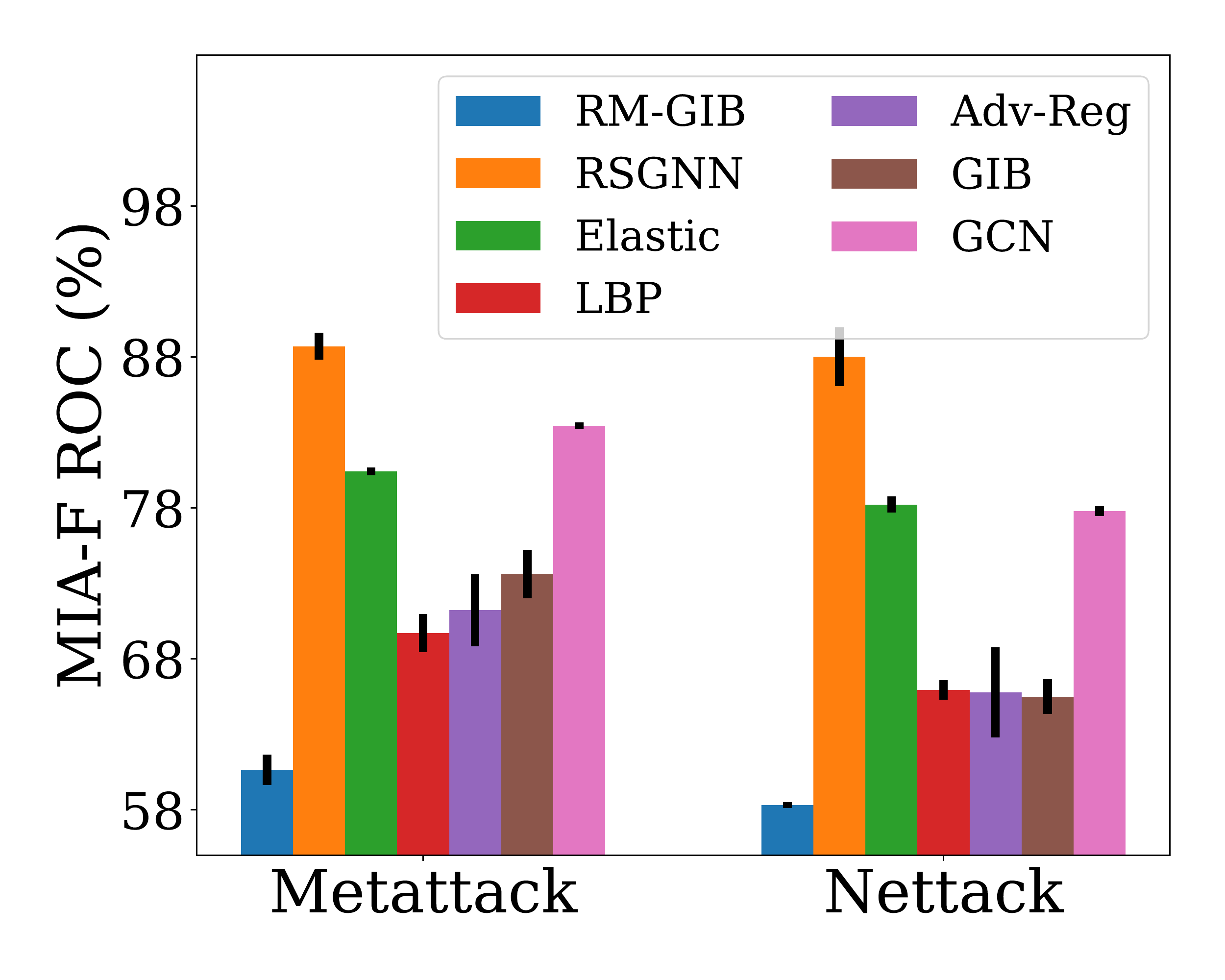}
        \vskip -1em
        \caption{MIA-F ROC on Pubmed}
    \end{subfigure}
    \vskip -1.5em
    \caption{Results on perturbed Cora and Pubmed graphs.}
    \vskip -1.5em
    \label{fig:attack_mia}
\end{figure}

\subsection{Results on Adverarially Perturbed Graphs} \label{sec:exp_adv}
To answer \textbf{RQ2}, we first compare {\method} with Robust GNNs on various perturbed graphs. Then, the performance of membership privacy-preserving on perturbed graphs is also evaluated.

\subsubsection{Robust Classification} Two types of adversarial attacks, i.e., Metattack and Nettack, are considered for all datasets. Metattack and Nettack will result in out of memory in attacking the large-scale dataset Flickr. Therefore, we only conduct experiments on Cora, Citeseer, and Pubmed. The detailed settings of attacks follow the description in Sec.~\ref{sec:exp_set}. The average results and standard deviations of 5 runs are reported in Tab.~\ref{tab:results_robust}, where we can observe:
\begin{itemize} [leftmargin=*]
    \item Our proposed {\method} achieves comparable/better results compared with the state-of-the-art robust GNNs on perturbed graphs, which indicates {\method} can mitigate the attribute noises and structural noises with the attribute and neighbor bottleneck. 
    \item Our {\method} performs much better than GIB, which also applies IB on graphs to filter out noises in attributes and structures. This is because self-supervision on neighbor bottleneck is adopted in {\method} to eliminate noisy neighbors irrelevant to label information. Meanwhile, incorporating pseudo labels of unlabeled nodes also benefits bottleneck code learning.
    \item On clean graphs, {\method} can also consistently outperform baselines including GCN. This is because clean graphs can contain superfluous information and inherent noises, which can be alleviated with the bottleneck in {\method}. 
\end{itemize}

\subsubsection{Membership Privacy Preserving}
 We also evaluate {\method} on perturbed graphs in terms of membership privacy-preserving. The most effective privacy-preserving baselines in Tab.~\ref{tab:results_mia} and robust GNNs in Tab.~\ref{tab:results_robust} are selected for comparison. The accuracy and MIA-F ROC on Pubmed and Cora that are perturbed by Metattack and Nettack are shown in Fig.~\ref{fig:attack_mia} and Fig.~\ref{app:fig_attack_mia}, respectively. From this figure, we can find that robust GNNs generally fail in preserving privacy. And privacy-preserving baselines give poor classification performance on perturbed graphs. In contrast, {\method} can simultaneously preserve membership privacy and give robust predictions in a unified framework.
\begin{figure}[t]
    \small
    \centering
    \begin{subfigure}{0.49\linewidth}
        \includegraphics[width=0.96\linewidth]{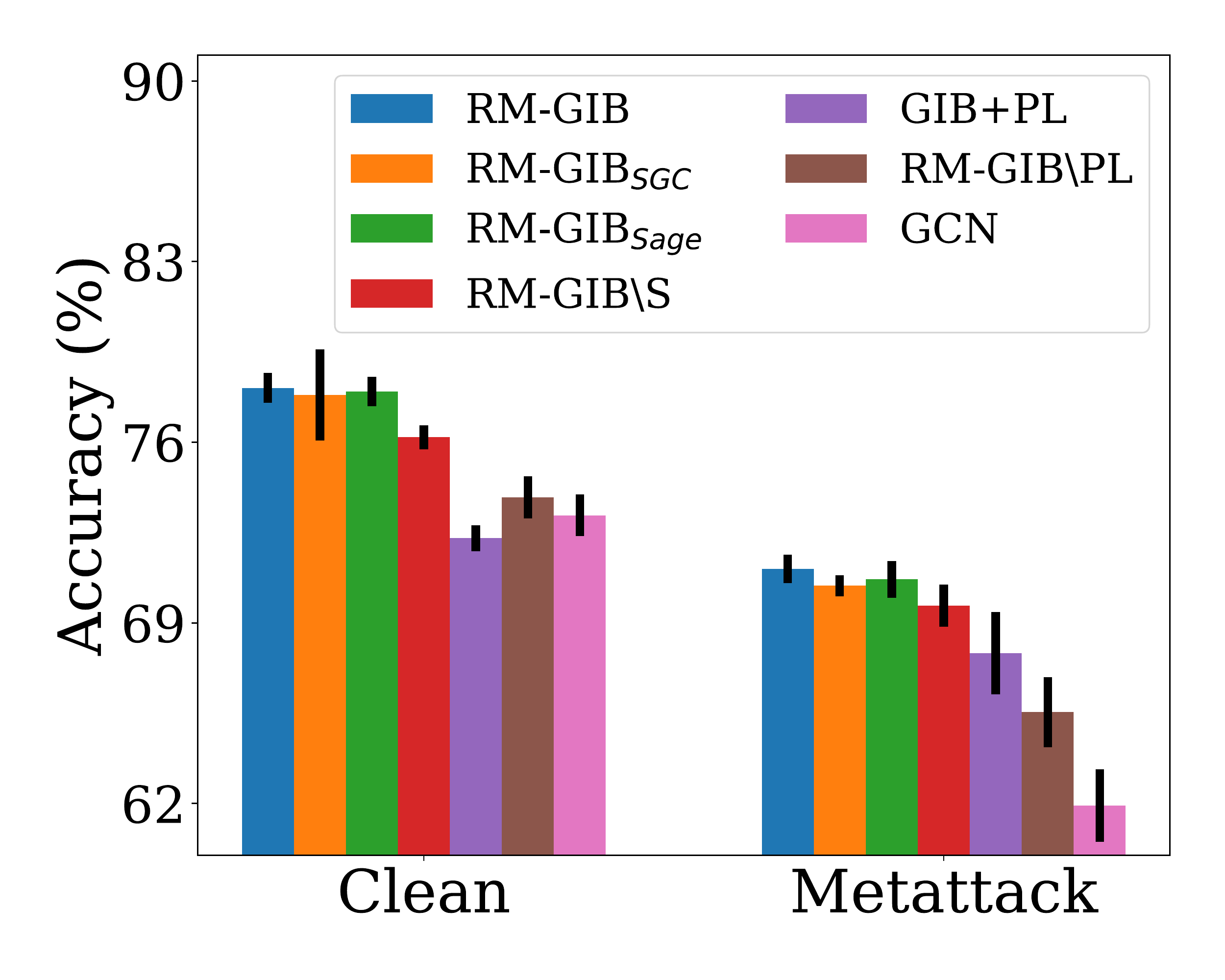}
        \vskip -1em
        \caption{Accuracy}
    \end{subfigure}
    \begin{subfigure}{0.49\linewidth}
        \includegraphics[width=0.96\linewidth]{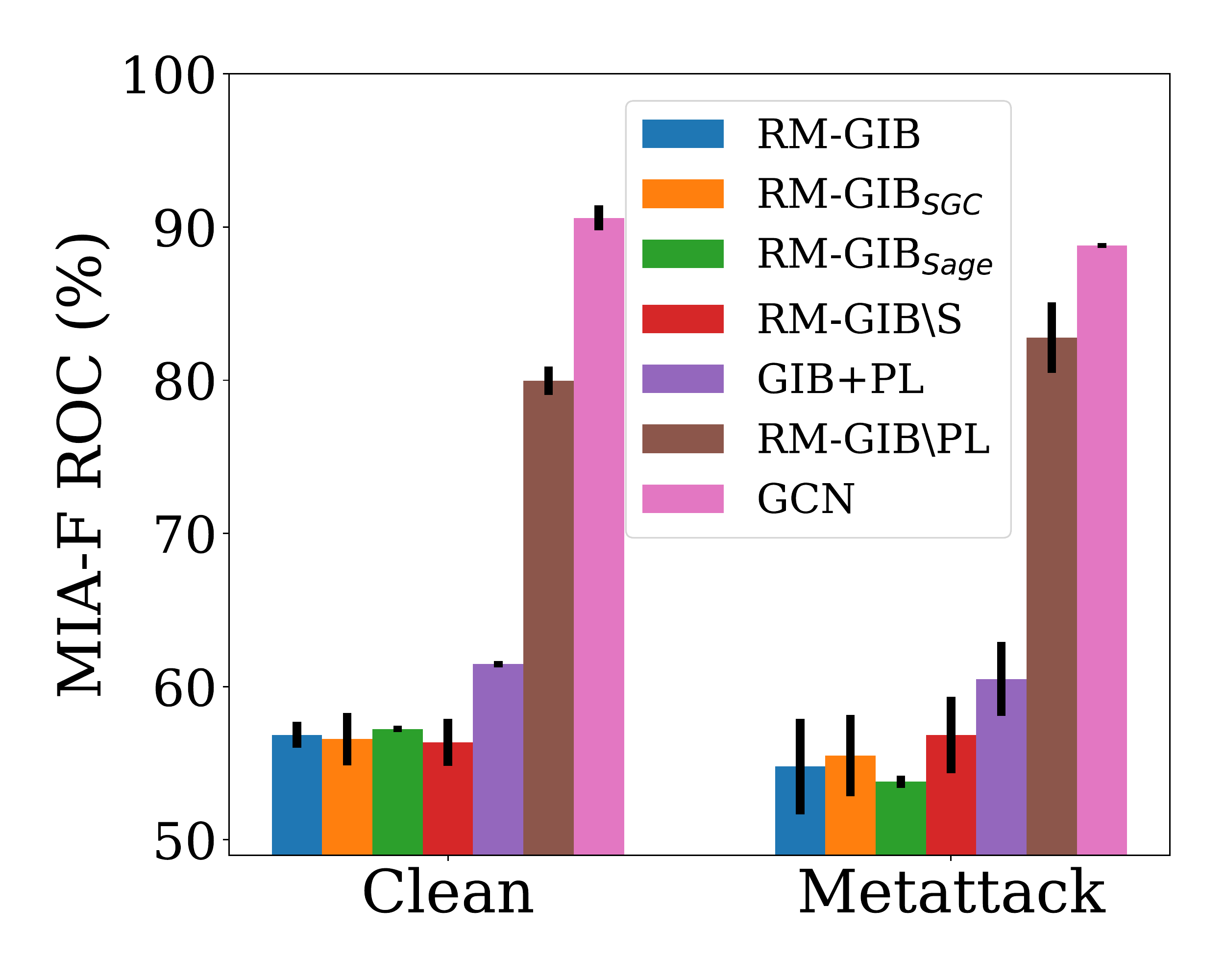}
        \vskip -1em
        \caption{MIA-F ROC}
    \end{subfigure}
    \vskip -1.5em
    \caption{Ablation studies on the Cora graph.}
    \vskip -1.5em
    \label{fig:abla}
\end{figure}

\subsection{Ablation Study}
To answer \textbf{RQ3}, we conduct an ablation study to understand the effects of the proposed graph information bottleneck, self-supervision on the neighbor bottleneck, and adoption of pseudo labels. To demonstrate the effectiveness of the self-supervision on the neighbor bottleneck, we set  $\gamma$ as 0 when we train {\method} and denote this variant as {\method}$\backslash$S. Moreover, to show our {\method} can better bottleneck noisy neighbors, a GIB+PL model which trains GIB~\cite{wu2020graph} with pseudo labeling is adopted as a reference. We train a variant {\method}$\backslash$PL that does not incorporate any pseudo labels of unlabeled nodes in the optimization to show the benefits of using pseudo labels in the training. To prove the flexibility of {\method}, we train two variants of {\method} that use SGC and GraphSage as the predictor, which correspond to {\method}$_{SGC}$ and {\method}$_{Sage}$. Results of classification and membership privacy-preserving on clean graphs and Metattack perturbed graphs are reported in Fig.~\ref{fig:abla}. We only show results on Cora as we have similar observations on other datasets. Concretely, we observe that:
\begin{itemize}[leftmargin=*]
    \item {\method}$_{SGC}$ and {\method}$_{Sage}$ achieve comparable results in both robustness and membership privacy-preserving, which shows the flexibility of our proposed {\method}.
    \item The accuracy of {\method}$\backslash$S and GIB+PL is worse than {\method} especially on perturbed graphs, which verifies self-supervision on neighbor bottleneck can benefit filtering out noisy neighbors.
    \item {\method} outperforms {\method}$\backslash$PL in both accuracy and membership privacy preserving. This shows the effectiveness of adopting pseudo labels to IB for preserving membership privacy. Pseudo labels on unlabeled nodes also improve the quality of the bottleneck code, resulting in better classification performance. 
\end{itemize}

\subsection{Hyperparameter Sensitivity Analysis} \label{sec:hyper}
\begin{figure}[t]
    \small
    \centering
    \begin{subfigure}{0.49\linewidth}
        \includegraphics[width=\linewidth]{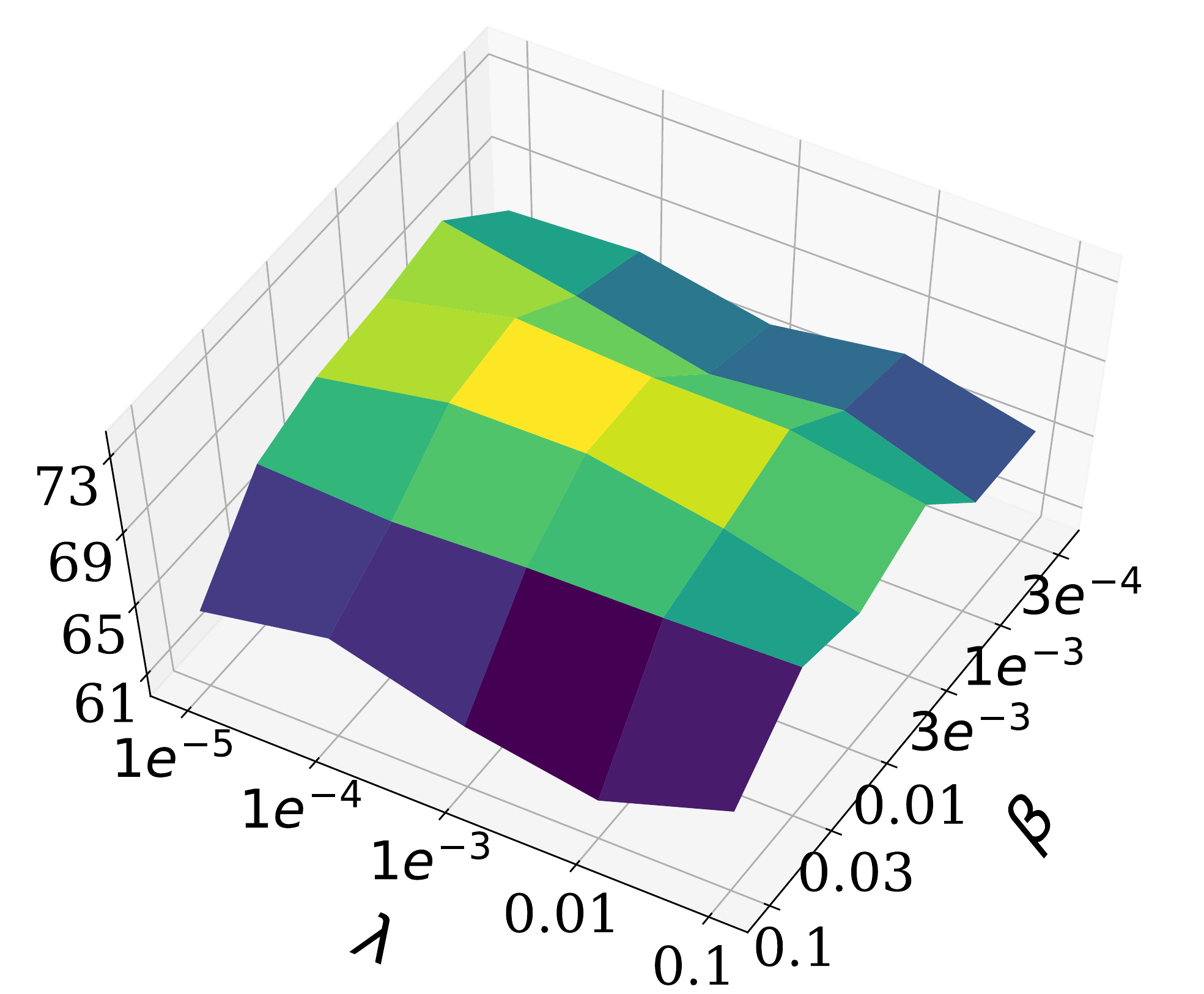}
        \vskip -0.5em
        \caption{Accuracy}
    \end{subfigure}
    \begin{subfigure}{0.49\linewidth}
        \includegraphics[width=\linewidth]{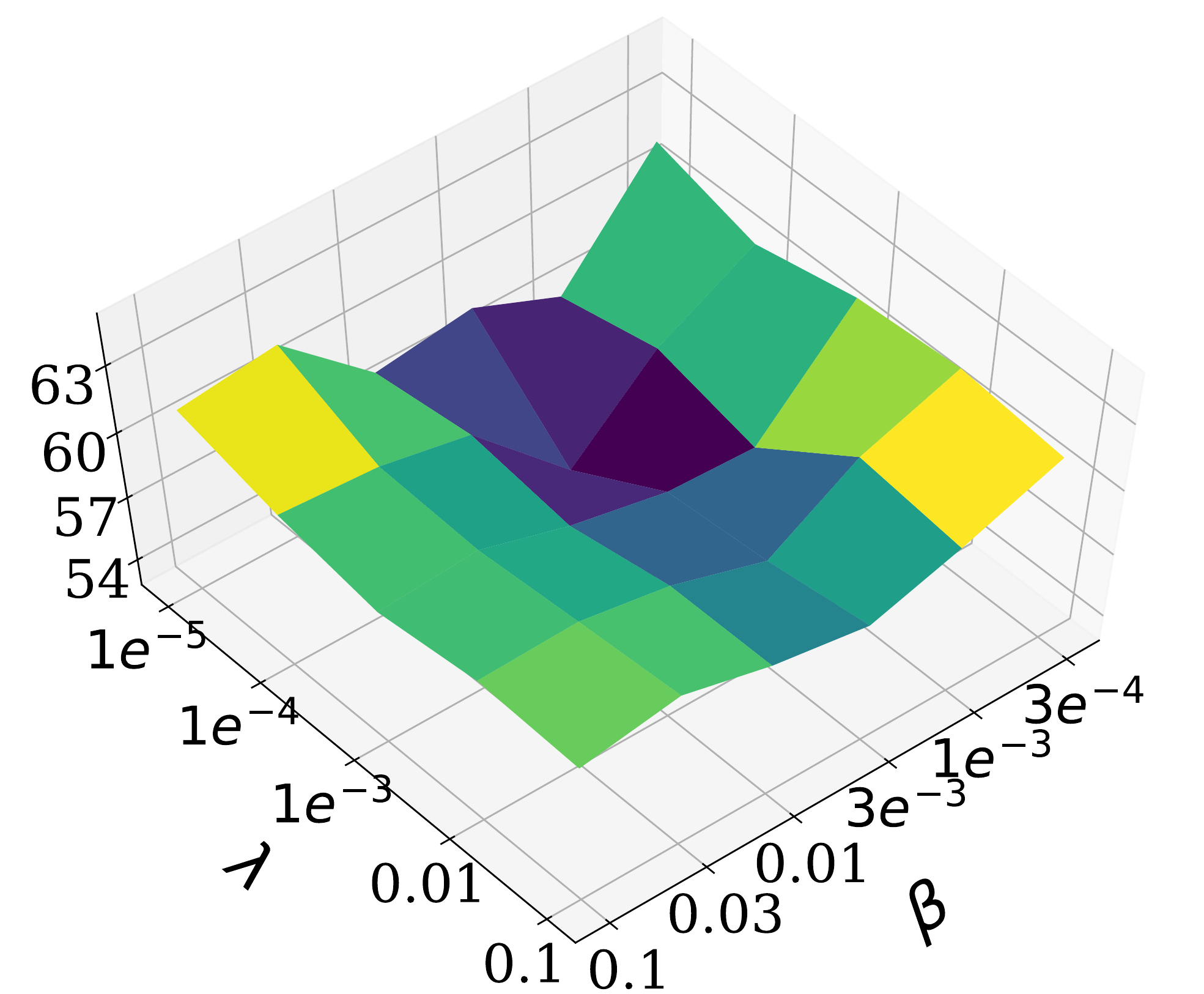}
        \vskip -0.5em
        \caption{MIA-F ROC}
    \end{subfigure}
    \vskip -1.3em
    \caption{Hyperparameter analysis on the perturbed Cora.}
    \vskip -1em
    \label{fig:hyper}
\end{figure}

In this subsection, we conduct hyperparameter sensitivity analysis to investigate how $\beta$ and $\gamma$ affect the {\method}, where $\beta$ controls the regularization on the bottleneck code and $\gamma$ controls the contributions of self-supervision on the neighbor bottleneck. More specifically, we vary $\beta$ and $\gamma$ as $\{0.0003, 0.0001, 0.003, 0.001, 0.03, 0.1 \}$ and $\{0.00001, 0.0001, 0.001, 0.01, 0.1\}$, respectively. We report the accuracy and MIA-F ROC on Cora graph perturbed by Metattack. Similar trends are also observed on other datasets and attack methods.
The results are shown in Fig.~\ref{fig:hyper}. We find that: (i) With the increase of $\beta$, the performance of classification and membership privacy-preserving both become better. This is because with very small $\beta$, the regularization will be too weak, which can cause overfitting and failure in filtering out noisy information. When $\beta$ is very large, the strong constraint will lead to poor generalization ability of bottleneck code, resulting in worse performance of both classification and membership privacy; (ii) With the increment of $\gamma$, the classification accuracy on perturbed graphs tends to first increase and decrease. And its effects on preserving membership privacy is negligible. When $\gamma$ is in $[0.0001, 0.001]$,  {\method} generally gives good classification performance. 

\section{Conclusion And Future Work}
In this paper, we study a novel problem of developing a unified framework that can simultaneously achieve robustness and preserve membership privacy. We verify that IB has potential to eliminate the noises and adversarial perturbations in the data. In addition, IB regularizes the predictions on labeled samples, which can benefit membership privacy. However, the deployment of IB on graph-structured data is challenged by structural noises and shortage of labels in node classification on graphs. To address these issues, we propose a novel graph information bottleneck framework that separately bottlenecks the attribute and neighbor information to handle attribute and structural noises. A self-supervision loss is applied to neighbor bottleneck to further help to filter out adversarial edges and inherent structural noises. Moreover, pseudo labels of unlabeled nodes are incorporated in optimization with pseudo labels to enhance membership privacy. There are two directions that need further investigation. In this work, we only focus on membership inference attacks. We will investigate whether IB can help defend against other privacy attacks such as attribute inference attacks. Since IB can extract minimal sufficient information, it would be interesting to investigate whether the sensitive information of users such as race can be removed for fairness. 

\section{Acknowledgement}
This material is based upon work partially supported by National Science Foundation (NSF) under grant number IIS-1909702 and the Army Research Office (ARO) under grant number W911NF21-1-0198 to Suhang Wang. The findings and conclusions in this paper do not necessarily reflect the view of the funding agency.
\newpage

\bibliographystyle{ACM-Reference-Format}
\bibliography{ref}
\appendix

\section{Dataset} \label{app:dataset}
Cora, Citeseer, and Pubmed are citation networks, where nodes in the graphs represent the papers and edges denote citation relationship. The attributes of the nodes are the bag-of-words of these papers. For small citation graphs, i.e., Cora and Citeseer, we randomly sample 2\% nodes as the training set. For the large citation graph Pubmed, we randomly sample 0.5\% nodes as the training set.  As for Flickr~\cite{zeng2020graphsaint}, it is a large-scale graph to categorize the type of images. Each node represents an image and the image description is used as a node attribute. Edges are formed between nodes sharing common properties. We randomly sample 2\% nodes from Flickr as the training set. Splits of validation and test sets on all datasets follow the cited papers for consistency. Note that the training node set doesn't overlap with the validation and test sets.

\section{Baselines} \label{app:baseline}
To evaluate the performance in preserving membership privacy, we compare {\method} with the following representative and state-of-the-art methods in defending membership inference attacks:
\begin{itemize}[leftmargin=*]
    \item GCN~\cite{kipf2016semi}: This is a representative graph convolutional network which defines graph convolution with spectral analysis.
    \item GCN+PL~\cite{lee2013pseudo}: A GCN is firstly trained to obtain pseudo labels. Then, pseudo labels of unlabeled nodes and labels of labeled nodes are used to retrain the GCN. 
    \item GIB~\cite{wu2020graph}: It proposes a graph information bottleneck that regularizes the structural and attribute information in GAT~\cite{velivckovic2017graph}.
    \item Adv-Reg~\cite{nasr2018machine}: Min-max game between the training model and the membership inference attacker is introduced as regularization for membership privacy-preserving.
    \item DP-SGD~\cite{abadi2016deep}: This is a differentially private mechanism that adds noises to gradients during optimization for preserving privacy.
    \item LBP~\cite{olatunji2021membership}: This is an output perturbation method by adding noise to the posterior before it is released to end users.
    \item NSD~\cite{olatunji2021membership}:  It randomly chooses neighbors of the queried node in inference to limit the amount of information used in the target model for membership privacy protection.
\end{itemize}
Apart from GCN and GIB, we also compare the following representative and state-of-the-art robust GNNs to evaluate the robustness of {\method} against adversarial attacks on graphs:
\begin{itemize} [leftmargin=*]
    \item GCN-jaccard~\cite{wu2019adversarial}: It preprocesses a graph by removing edges linking nodes with low Jaccard feature similarity, then trains a GCN on the preprocessed graph.
    \item GCN-SVD~\cite{entezari2020all}: It uses a low-rank approximation of the perturbed graph to defend against graph adversarial attacks with the observation that adversarial edges often result in a high-rank adjacency matrix.
    \item Elastic~\cite{liu2021elastic}: Elastic designs a robust message-passing mechanism which incorporates $l_1$-based graph smoothing in GNNs. 
    \item RSGNN~\cite{dai2022towards}: This is a state-of-the-art robust GNN that denoises and densifies the noisy graph to give robust predictions.
\end{itemize}

\begin{table*}[t]
    \small
    \centering
    \caption{Results of defending membership inference attack (Accuracy(\%)$\uparrow$ | MIA-F ROC(\%) $\downarrow$) with various label rates.}
    \vskip -1em
    \begin{tabularx}{0.92\textwidth}{p{0.1\textwidth}p{0.1\textwidth}CCCC}
    \toprule
    Dataset & Method & 2\% & 4\% & 6\% & 8\% \\
    \midrule
    \multirow{3}{*}{Cora}
    & GCN & 73.2$\pm$0.8 | 89.4$\pm$0.5 & 79.4$\pm$0.2 | 81.3$\pm$1.4 & 81.2$\pm$0.2 | 78.0$\pm$0.4 & \textbf{82.1$\pm$0.3} | 74.3$\pm$0.1 \\
    & GIB & 72.5$\pm$0.7 | 86.6$\pm$0.8 & 78.8$\pm$0.5 | 78.6$\pm$0.7 & 80.6$\pm$1.5 | 71.4$\pm$1.8 & 80.9$\pm$0.8 | 67.8$\pm$0.6 \\
    & RM-GIB & \textbf{78.5$\pm$0.6} | \textbf{56.4$\pm$0.2} & \textbf{79.6$\pm$0.6} | \textbf{56.9$\pm$0.3} & \textbf{81.9$\pm$0.4} | \textbf{55.9$\pm$0.6} & 81.9$\pm$0.3 | \textbf{54.4$\pm$1.0} \\
    \midrule
    \multirow{3}{*}{Citeseer}
    & GCN & 70.2$\pm$0.2 | 88.5$\pm$1.8 & 71.3$\pm$0.4 | 83.1$\pm$0.3 & 73.6$\pm$0.1 | 76.0$\pm$0.3 & 73.9$\pm$0.1 | 73.2$\pm$0.1 \\
    & GIB & 70.1$\pm$1.1 | 87.4$\pm$0.6 & 72.1$\pm$0.6 | 80.4$\pm$2.1 & 74.8$\pm$0.5 | 70.9$\pm$0.3 & 74.6$\pm$0.7 | 69.9$\pm$0.9 \\
    & RM-GIB & \textbf{73.9$\pm$0.6} | \textbf{55.2$\pm$0.8} & \textbf{73.6$\pm$0.8} | \textbf{53.0$\pm$0.1} & \textbf{76.1$\pm$0.3} | \textbf{50.3$\pm$0.7} & \textbf{76.4$\pm$0.7} | \textbf{50.2$\pm$1.8}\\
    \midrule
    \multirow{3}{*}{Pubmed}
    & GCN & 81.0$\pm$0.1 | 56.6$\pm$0.1 & 82.8$\pm$0.4 | 56.6$\pm$0.1 & 83.9$\pm$0.1 | 54.9$\pm$0.1 & 85.3$\pm$0.1 | 53.0$\pm$0.1 \\
    & GIB & 81.9$\pm$0.1 | 56.1$\pm$0.2 & 84.0$\pm$0.2 | 53.7$\pm$0.4 & 85.1$\pm$0.3 | 52.0$\pm$0.1 & 85.5$\pm$0.8 | 51.3$\pm$0.1 \\
    & RM-GIB & \textbf{84.0$\pm$0.1} | \textbf{50.3$\pm$0.5} & \textbf{85.2$\pm$0.4} | \textbf{49.8$\pm$0.7} & \textbf{85.9$\pm$0.3} | \textbf{50.1$\pm$0.3} & \textbf{86.4$\pm$0.2} | \textbf{50.1$\pm$0.1} \\
    \bottomrule
    \end{tabularx}
    \label{tab:app_mem}
    % \vskip -1.2em
\end{table*}

\section{Implementation Details}~\label{app:imple}
A 2-layer MLP is deployed as the attribute bottleneck. The neighbor bottleneck also uses a 2-layer MLP. As for the predictor, we use a 2-layer GCN without on default. The mutual information estimator used for self-supervision on neighbor bottleneck also deploys a 2-layer MLP. All the hidden dimensions of the neural networks are set as 256. For the hyperparameter $T$ which is the threshold to determine the negative neighbors for self-supervision, it is set as 0.5 for all experiments. As for the hyperparameters $\beta$ and $\gamma$ used in the final objective function Eq.(\ref{eq:final}), they are selected based on accuracy on the validation set with grid search. Specifically, we vary $\beta$ and $\gamma$ as $\{0.0003, 0.0001, 0.003, 0.001, 0.03, 0.1 \}$ and $\{0.00001, 0.0001, 0.001, 0.01, 0.1\}$, respectively.

\section{Proof Details}~\label{app:proof}
Recall that in IB, for a given training sample $(\mathbf{x}_n, y_n)$, its distribution of $\mathbf{z}$ is obtained by $P(\mathbf{z}|\mathbf{x}_n, y_n;\theta)=f_{\theta}(\mathbf{z}, \mathbf{x}_n)$, where $f_{\theta}(\mathbf{z}, \mathbf{x}_n)$ is the probability density function modeled by the nerual network with parameters $\theta$.
In the practice of computing mutual information, $P(\mathbf{x},y,\mathbf{z}; \theta)$ is approximated with the empirical data distribution $P(\mathbf{x},y,\mathbf{z}; \theta)=\frac{1}{N} \sum_{n=1}^N \delta_{\mathbf{x}_n}(\mathbf{x}) \delta_{y_n}(y) f_{\theta}(\mathbf{z}, \mathbf{x}_n)$, where $\delta()$ is the Dirac function. Then, we can have the following equations:

\begin{equation}
    P(\mathbf{x}, \mathbf{z}; \theta) = \int_{y} P(\mathbf{x},y,\mathbf{z}; \theta) dY=\frac{1}{N} \sum_{n=1}^N \delta_{\mathbf{x}_n}(\mathbf{x})f_{\theta}(\mathbf{z}, \mathbf{x}_n)
\end{equation}
\begin{equation}
    P(\mathbf{x},y;\theta) = \int_{\mathbf{z}} P(\mathbf{x},y,\mathbf{z};\theta) d\mathbf{z}=\frac{1}{N} \sum_{n=1}^N \delta_{\mathbf{x}_n}(\mathbf{x}) \delta_{y_n}(y)
\end{equation}
\begin{equation}
    P(\mathbf{x}; \theta) = \int_{y} P(\mathbf{x},y;\theta) d\mathbf{x}=\frac{1}{N} \sum_{n=1}^N \delta_{\mathbf{x}_n}(\mathbf{x})
\end{equation}

The $I_{\theta}(\mathbf{z};y|\mathbf{x})$ can be computed by:

\begin{equation}
\begin{aligned}
& I_{\theta}(\mathbf{z};y|\mathbf{x})  \\
&= \int_{\mathbf{x}} \int_{y} \int_{\mathbf{z}} p(\mathbf{x},y,\mathbf{z}; \theta) \log \frac{P(\mathbf{x}; \theta)P(\mathbf{x},y,\mathbf{z};\theta)}{P(\mathbf{x},y;\theta)P(\mathbf{x},\mathbf{z};\theta)} d\mathbf{x}dyd\mathbf{z}\\
& = \frac{1}{N} \int_{\mathbf{x}} \int_{y} \int_{\mathbf{z}} \Big( \sum_{n=1}^N \delta_{\mathbf{x}_n}(\mathbf{x}) \delta_{y_n}(y) f_{\theta}(\mathbf{z}, \mathbf{x}_n) \Big) \\
& \cdot \log \frac{\big(\sum_{n=1}^N \delta_{\mathbf{x}_n}(\mathbf{x}) \big) \cdot \big(\sum_{n=1}^N \delta_{\mathbf{x}_n}(\mathbf{x}) \delta_{y_n}(y) f_{\theta}(\mathbf{z}, \mathbf{x}_n)\big)}{\big( \sum_{n=1}^N \delta_{\mathbf{x}_n}(\mathbf{x}) \delta_{y_n}(y) \big) \cdot \big(\sum_{n=1}^N \delta_{\mathbf{x}_n}(\mathbf{x}) f_{\theta}(\mathbf{z}, \mathbf{x}_n)  \big)} d\mathbf{x}dyd\mathbf{z}\\
& = \frac{1}{N} \int_{\mathbf{z}} \sum_{n=1}^N f_{\theta}(\mathbf{z}, \mathbf{x}_n) \log \frac{f_{\theta}(\mathbf{z}, \mathbf{x}_n)}{f_{\theta}(\mathbf{z}, \mathbf{x}_n)}  = 0
\end{aligned}
\end{equation}
Based on the above proof, we verify that $I_{\theta}(\mathbf{z};y|\mathbf{x})=0$ regardless the value of model parameters. Thus, we can derive the first line of Eq.(\ref{eq:pre_mem}) in our paper.

\section{Time Complexity Analysis}
We analyze the time complexity of the proposed RM-GIB in the following. The time complexity mainly comes from the pretraining of self-supervisor for the neighbor bottleneck, and the training of RM-GIB. Let $h$ and $K$ denote the embedding dimension and training epochs, respectively. The cost of training the self-supervisor is approximately $O(Khd|\mathcal{V}|)$, where $d$ is the average degree of nodes and $|\mathcal{V}|$ is the number of nodes in the graph. Next, we analyze the time complexity of the optimization of RM-GIB. The time complexity of attribute bottleneck and neighbor bottleneck in each epoch are $O(h|\mathcal{V}|)$ and $O(hd|\mathcal{V}|)$, respectively.  As for the computation cost of the predictor is approximately $O(hd|\mathcal{V}|)$ in each epoch. The privacy-preserving optimization requires firstly training RM-GIB for pseudo label collection followed by the optimization on the enlarged label set. Hence, the time complexity of optimizing RM-GIB is $O(2Kh(2d+1)|\mathcal{V}|)$. Combining the training of self-supervisor, the overall time complexity for training is $O(Kh(4d+3)|\mathcal{V}|)$. Our {\method} is linear to the size of the graph, which proves its scalability.

\section{Additional Experimental Results}

\begin{figure}[h!]
    \small
    \centering
    \vskip -1em
    \begin{subfigure}{0.49\linewidth}
        \includegraphics[width=0.98\linewidth]{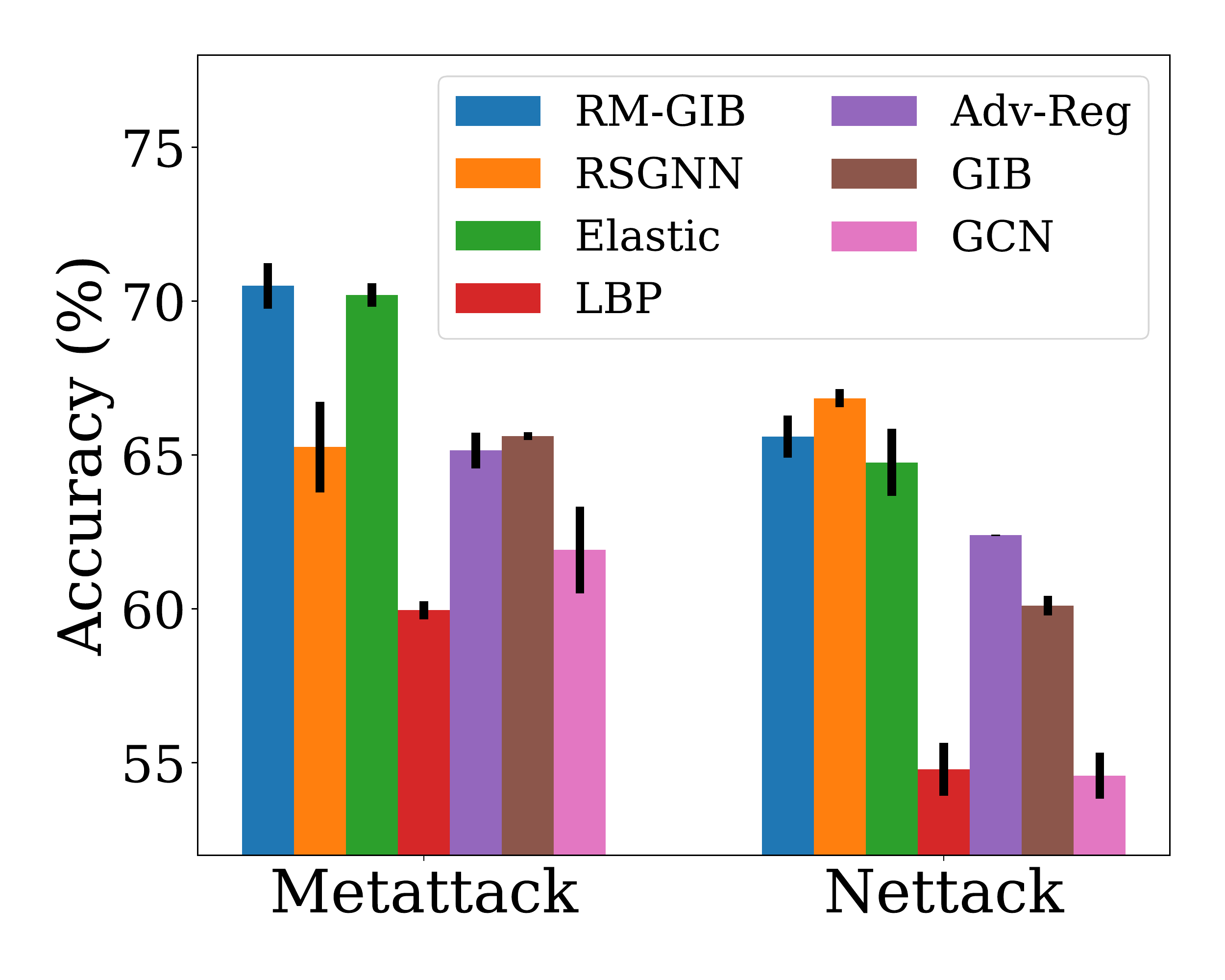}
        \vskip -1em
        \caption{Accuracy on Cora}
    \end{subfigure}
    \begin{subfigure}{0.49\linewidth}
        \includegraphics[width=0.98\linewidth]{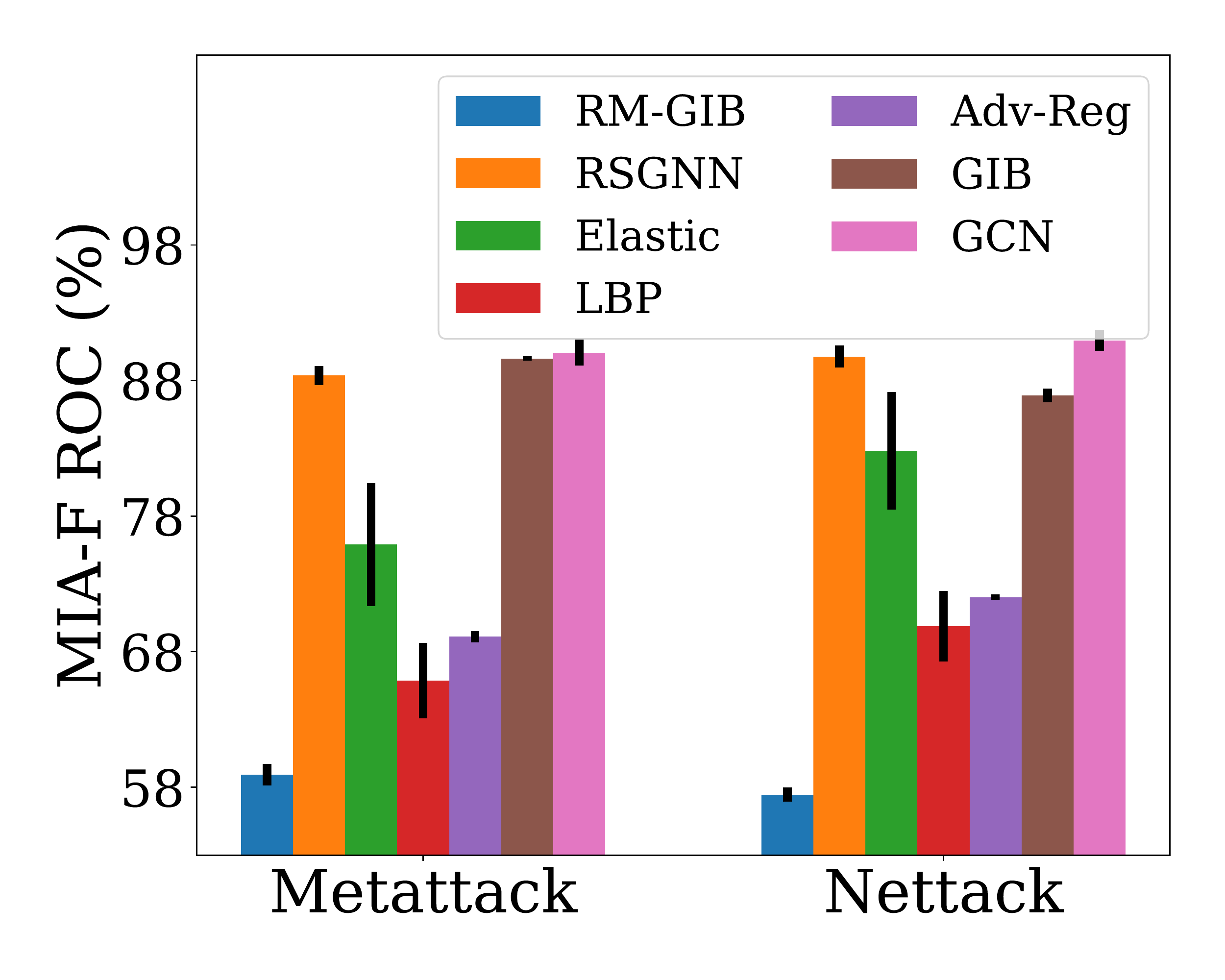}
        \vskip -1em
        \caption{MIA-F ROC on Cora}
    \end{subfigure}
    \vskip -1em
    \caption{Additional results on the perturbed Cora.}
    \vskip -1em
    \label{app:fig_attack_mia}
\end{figure}
The additional results on the perturbed Cora graph are shown in Fig.~\ref{app:fig_attack_mia}, which have the same observations as Fig.~\ref{fig:attack_mia}.

\noindent \textbf{Impacts of Label Rates.} \label{app:add_label}
We add the experiments that vary label rates by \{2\%, 4\%, 6\%, 8\%\} to verify our motivation and the effectiveness of our RM-GIB. All the hyperparameters of GCN, GIB, and our RM-GIB are tuned on the validation set for a fair comparison. The results are presented in Table~\ref{tab:app_mem}. We can observe that:
\begin{itemize}[leftmargin=*]
    \item When the label rates are small, GIB gives high MIA-F ROC scores and marginally outperforms GCN in privacy preservation.  This verifies that GIB is vulnerable to membership inference attack under a semi-supervised learning setting.
    \item Our method RM-GIB can consistently achieve a very low MIA-F ROC score (close to 50\%) with different sizes of labeled nodes. This demonstrates the effectiveness of our RM-GIB in membership privacy preservation under different data settings.
\end{itemize}

\begin{table}[t]
    \small
    \centering
    \caption{Impacts of labels rates in defending metattack.}
    \vskip -1em
    \begin{tabular}{llcccc}
    \toprule
    & & 2\% & 4\% & 6\% & 8\%\\
    \midrule
    \multirow{3}{*}{Cora}
    & GCN & 62.7$\pm$0.6 & 71.9$\pm$0.2 & 76.0$\pm$0.2 & 77.7$\pm$0.3 \\
    & GIB & 65.6$\pm$0.1 & 74.0$\pm$0.7 & 77.5$\pm$1.0 & 78.4$\pm$0.5 \\
    & RM-GIB & \textbf{71.1$\pm$0.6} & \textbf{75.7$\pm$0.6} & \textbf{78.4$\pm$0.5} & \textbf{79.6$\pm$0.6} \\
    \midrule
    \multirow{3}{*}{Citeseer} 
    & GCN & 66.1$\pm$0.5 & 67.9$\pm$1.9 & 68.3$\pm$0.8 & 71.1$\pm$0.4 \\
    & GIB & 66.8$\pm$0.7 & 68.90.7 & 69.4$\pm$0.2 & 72.2$\pm$0.5 \\
    & RM-GIB & \textbf{72.1$\pm$0.9} & \textbf{71.9$\pm$0.9} & \textbf{74.5$\pm$0.9} & \textbf{74.6$\pm$0.3} \\
    \midrule
    \multirow{3}{*}{Pubmed}
    & GCN & 70.3$\pm$0.1 & 72.1$\pm$0.1 & 72.9$\pm$0.1 & 74.1$\pm$0.3 \\
    & GIB & 70.8$\pm$0.2 & 73.4$\pm$0.2 & 74.4$\pm$0.2 & 75.2$\pm$0.3 \\
    & RM-GIB & \textbf{81.2$\pm$0.2} & \textbf{81.9$\pm$0.3} & \textbf{83.1$\pm$0.5} & \textbf{84.4$\pm$0.6} \\
    \bottomrule
    \end{tabular}
    % \vskip -1em
    \label{tab:app_meta}
\end{table}

We also show the accuracy (\%)) of defending metattack (20\% perturbation rate) under different label rates in Tab.~\ref{tab:app_meta}. Our RM-GIB consistently performs better than GIB by a large margin in defending graph adversarial attacks given different sizes of labels.

\begin{table}[t]
    \small
    \centering
    \caption{Results (\%) of varying pseudo label Sizes.}
    \vskip -1em
    \begin{tabular}{lccccc}
    \toprule
    & 5\% & 10\% & 20\% & 50\% & 100\%\\
    \midrule
    MIA-F ROC & 68.0$\pm$0.9 & 63.2$\pm$3.3 & 62.0$\pm$4.3 & 58.1$\pm$2.1 & 54.8$\pm$3.1\\
    Accuracy  & 68.1$\pm$0.5 & 69.8$\pm$1.2 & 70.5$\pm$0.3 & 71.1$\pm$0.6 & 71.7$\pm$0.4 \\
    \bottomrule
    \end{tabular}
    % \vskip -1em
    \label{tab:app_pse_size}
\end{table}
\noindent \textbf{Varying Sizes of Pseudo Labels.} We vary the rates of unlabeled nodes used for the pseudo-label generation by \{5\%, 10\%, 20\%, 50\%, 100\%\} . Experiments are  conducted on the Cora graph. For the adversarial attacks, we apply metattack with 20\% perturbation rate. All other settings are the same as the description in Sec.~\ref{sec:exp_set}. The results are shown in Tab.~\ref{tab:app_pse_size}.
We can observe from the results that with the increase of pseudo labels, the performance in defending membership inference attack and adversarial attacks will both increase. This demonstrates the effectiveness of incorporating pseudo labels. It justifies that we should generate pseudo labels for all the unlabeled nodes in the graph.

\begin{table}[t]
    \small
    \centering
    \caption{Accuracy on attribute-perturbed only graphs.}
    \vskip -1em
    \begin{tabularx}{0.9\linewidth}{lCCC}
    \toprule
    Dataset & GCN & GIB & {\method}\\
    \midrule
    Cora & 70.3$\pm$1.3 & 74.3$\pm$0.2 & \textbf{78.2$\pm$0.7} \\
    Citeseer & 70.7$\pm$0.5 & 71.6$\pm$0.2 & \textbf{73.9$\pm$0.9} \\
    \bottomrule
    \end{tabularx}
    \vskip -1em
    \label{tab:app_meta_attr}
\end{table}

\noindent \textbf{Results on Attribute Perturbation.}
we conduct experiments on attribute-perturbed only graphs to empirically verify the effectiveness of our methods in defending against noises in attributes. We apply metattack to poison the attributes of the Cora and Citeseer graphs with the perturbation rate set as 20\%. The other settings are the same as the description in Sec.~\ref{sec:exp_set}. The results are shown in Tab.~\ref{tab:app_meta_attr}, where we can observe that our RM-GIB performs better than GIB on attribute-perturbed graphs. This verifies the effectiveness of our method in defending noises in node attributes.

\end{document}